\documentclass[runningheads]{llncs}
\usepackage{graphicx}
\usepackage{amsmath,amssymb} 
\usepackage{color}
\usepackage[width=122mm,left=12mm,paperwidth=146mm,height=193mm,top=12mm,paperheight=217mm]{geometry}
\usepackage{algpseudocode}
\usepackage{algorithm}
\usepackage{booktabs}
\usepackage{array}

\usepackage{bm}
\usepackage{dsfont}
\usepackage{placeins}
\usepackage[table,xcdraw]{xcolor}
\DeclareMathOperator*{\argmin}{\arg\!\min}

\usepackage{caption}
\usepackage{subcaption}
\let\llncssubparagraph\subparagraph
\let\subparagraph\paragraph
\usepackage[compact]{titlesec}
\let\subparagraph\llncssubparagraph

\titlespacing\section{0pt}{12pt plus 4pt minus 2pt}{5pt plus 2pt minus 2pt}
\titlespacing\subsection{0pt}{12pt plus 4pt minus 2pt}{3pt plus 2pt minus 2pt}
\titlespacing\subsubsection{0pt}{6pt plus 4pt minus 2pt}{2pt plus 2pt minus 2pt}

\expandafter\def\expandafter\normalsize\expandafter{%
    \normalsize
    \setlength\abovedisplayskip{5pt}
    \setlength\belowdisplayskip{5pt}
    \setlength\abovedisplayshortskip{5pt}
    \setlength\belowdisplayshortskip{5pt}
}

\makeatletter
\newcommand{\thickhline}{%
    \noalign {\ifnum 0=`}\fi \hrule height 1pt
    \futurelet \reserved@a \@xhline
}
\newcolumntype{"}{@{\hskip\tabcolsep\vrule width 1pt\hskip\tabcolsep}}
\makeatother

\begin{document}
\pagestyle{headings}
\mainmatter

\title{Efficient Continuous Relaxations for Dense CRF} 

\titlerunning{Efficient Continuous Relaxations for Dense CRF}
\newcommand*\samethanks[1][\value{footnote}]{\footnotemark[#1]}

\authorrunning{Desmaison, Bunel, Kohli, Torr, Kumar}

\author{Alban Desmaison\thanks{Joint first authors}\textsuperscript{1} \and Rudy Bunel\samethanks\textsuperscript{1} \and Pushmeet Kohli\textsuperscript{2} \and Philip H.S. Torr\textsuperscript{1} \and \mbox{M. Pawan Kumar\textsuperscript{1}}}
\institute{Department of Engineering Science, University of Oxford \mbox{\email{ \{alban, rudy, pawan\}@robots.ox.ac.uk, philip.torr@eng.ox.ac.uk}}
 \and Microsoft Research \\\email{pkohli@microsoft.com}}

\maketitle

\begin{abstract}
Dense conditional random fields (CRF) with Gaussian pairwise potentials have emerged as a popular framework for several computer vision applications such as stereo correspondence and semantic segmentation.
By modeling long-range interactions, dense CRFs provide a more detailed labelling compared to their sparse counterparts.
Variational inference in these dense models is performed using a filtering-based mean-field algorithm in order to obtain a fully-factorized distribution minimising the Kullback-Leibler divergence to the true distribution.
In contrast to the continuous relaxation-based energy minimisation algorithms used for sparse CRFs, the mean-field algorithm fails to provide strong theoretical guarantees on the quality of its solutions.
To address this deficiency, we show that it is possible to use the same filtering approach to speed-up the optimisation of several continuous relaxations.
Specifically, we solve a convex quadratic programming (QP) relaxation using the efficient Frank-Wolfe algorithm.
This also allows us to solve difference-of-convex relaxations
via the iterative concave-convex procedure where each iteration requires solving a convex QP.
Finally, we develop a novel divide-and-conquer method to compute the subgradients of a linear programming relaxation that provides the best theoretical bounds for energy minimisation.
We demonstrate the advantage of continuous relaxations over the widely used mean-field algorithm on publicly available datasets.
\end{abstract}

\section{Introduction}
Discrete pairwise conditional random fields (CRFs) are a popular framework for modelling several problems in computer vision.
In order to use them in practice, one requires an energy minimisation algorithm that obtains the most likely output for a given input.
The energy function consists of a sum of two types of terms: unary potentials that depend on the label for one random variable at a time and pairwise potentials that depend on the labels of two random variables.

Traditionally, computer vision methods have employed sparse connectivity structures, such as
4 or 8 connected grid CRFs. Their popularity lead to a considerable research
effort in efficient energy minimisation algorithms. One of the biggest successes
of this effort was the development of several accurate continuous relaxations of
the underlying discrete optimisation problem~\cite{ravikumar,kleinbergTardos2002}. An important advantage of
such relaxations is that they lend themselves easily to analysis, which allows
us to compare them theoretically~\cite{kumar_relax_comp}, as well as establish bounds on the
quality of their solutions~\cite{chekuri2001approximation}.

Recently, the influential work of Kr\"{a}henb\"{u}hl and Koltun~\cite{densecrf-kra} has popularised
the use of dense CRFs, where each pair of random variables is connected by an
edge. Dense CRFs capture useful long-range interactions thereby providing finer
details on the labelling. However, modeling long-range interactions comes at the
cost of a significant increase in the complexity of energy minimisation. In order
to operationalise dense CRFs, Kr\"{a}henb\"{u}hl and Koltun~\cite{densecrf-kra} made two key
observations. First, the pairwise potentials used in computer vision typically
encourage smooth labelling. This enabled them to restrict themselves to the
special case of Gaussian pairwise potentials introduced by Tappen et al.~\cite{tappenGMRF}. Second, for this special case, it
is possible to obtain a labelling efficiently by using the mean-field
algorithm~\cite{kollerbook}. Specifically, the message computation required at each iteration
of mean-field can be carried out in $O(N)$ operations where $N$ is the number of
random variables (of the order of hundreds of thousands).
This is in contrast to a na\"ive implementation that requires $O(N^2)$ operations. The significant speed-up is
made possible by the fact that the messages can be computed using the filtering
approach of Adams et al. \cite{permuto}.

While the mean-field algorithm does not provide any theoretical guarantees on
the energy of the solutions, the use of a richer model, namely
dense CRFs, still allows us to obtain a significant improvement in the accuracy
of several computer vision applications compared to sparse CRFs~\cite{densecrf-kra}.
However, this still leaves open the intriguing possibility that the same
filtering approach that enabled the efficient mean-field algorithm can also be
used to speed-up energy minimisation algorithms based on continuous relaxations.
In this work, we show that this is indeed possible.

In more detail, we make three contributions to the problem of energy
minimisation in dense CRFs. First, we show that the conditional gradient of
a convex quadratic programming (QP) relaxation~\cite{ravikumar} can be computed in $O(N)$
complexity. Together with our observation that the optimal step-size of a descent
direction can be computed analytically, this allows us to minimise the QP
relaxation efficiently using the Frank-Wolfe algorithm~\cite{frank-wolfe}. Second, we show
that difference-of-convex (DC) relaxations of the energy minimisation problem can
be optimised efficiently using an iterative concave-convex procedure (CCCP).
Each iteration of CCCP requires solving a convex QP, for which we can once
again employ the Frank-Wolfe algorithm. Third, we show that a linear
programming (LP) relaxation~\cite{kleinbergTardos2002} of the energy minimisation problem can
be optimised efficiently via subgradient descent. Specifically, we design a
novel divide-and-conquer method to compute the subgradient of the LP. Each
subproblem of our method requires one call to the filtering approach. This
results in an overall run-time of $O(N\log(N))$ per iteration as opposed to
an $O(N^2)$ complexity of a na\"ive implementation. It is worth noting that the
LP relaxation is known to provide the best theoretical bounds for energy
minimisation with metric pairwise potentials~\cite{kleinbergTardos2002}.

Using standard publicly available datasets, we demonstrate the
efficacy of our continuous relaxations by comparing them to the widely used
mean-field baseline for dense CRFs.

\section{Related works}
Kr\"{a}henb\"{u}hl and Koltun popularised the use of densely connected CRFs at the pixel level~\cite{densecrf-kra}, resulting in significant improvements both in terms of the quantitative performance and in terms of the visual quality of their results.
By restricting themselves to Gaussian edge potentials, they made the computation of the message in parallel mean-field feasible.
This was achieved by formulating message computation as a convolution in a higher-dimensional space, which enabled the use of an efficient filter-based method~\cite{permuto}.

While the original work~\cite{densecrf-kra} used a version of mean-field that is not guaranteed to converge, their follow-up paper~\cite{denseccp_kra} proposed a convergent mean-field algorithm for negative semi-definite label compatibility functions.
Recently, Baqu{\'{e}} et al.~\cite{princip-parallel} presented a new algorithm that has convergence guarantees in the general case.
Vineet et al.~\cite{vibhav-high-order} extended the mean-field model to allow the addition of higher-order terms on top of the dense pairwise potentials, enabling the use of co-occurence potentials~\cite{ladicky2010graph} and $P^n$-Potts models~\cite{kohli2007p3}.

The success of the inference algorithms naturally lead to research in learning the parameters of dense CRFs.
Combining them with Fully Convolutional Neural Networks~\cite{long2015fully} has resulted in high performance on semantic segmentation applications~\cite{deeplab}.
Several works~\cite{learn_cnn,crfasrnn} showed independently how to jointly learn the parameters of the unary and pairwise potentials of the CRF.
These methods led to significant improvements on various computer vision applications, by increasing the quality of the energy function to be minimised by mean-field.

Independently from the mean-field work, Zhang et al.~\cite{denseQP} designed a different set of constraints that lends itself to a QP relaxation of the original problem.
Their approach is similar to ours in that they use continuous relaxation to approximate the solution of the original problem but differ in the form of the pairwise potentials.
The algorithm they propose to solve the QP relaxation has linearithmic complexity while ours is linear.
Furthermore, it is not clear whether their approach can be easily generalised to tighter relaxations such as the LP.

Wang et al. \cite{Wang_2015_CVPR} derived a semi-definite programming relaxation of the energy minimisation problem, allowing them to reach better energy than mean-field.
Their approach has the advantage of not being restricted to Gaussian pairwise potentials.
The inference is made feasible by performing low-rank approximation of the Gram matrix of the kernel, instead of using the filter-based method.
However, while the complexity of their algorithm is the same as our QP or DC relaxation, the runtime is significantly higher.
Furthermore, while the SDP relaxation has been shown to be accurate for repulsive pairwise potentials (encouraging neighbouring variables to take different labels)~\cite{goemansjacm95}, our LP relaxation provides the best guarantees for attractive pairwise potentials~\cite{kleinbergTardos2002}.

In this paper, we use the same filter-based method~\cite{permuto} as the one employed in mean-field.
We build on it to solve continuous relaxations of the original problem that have both convergence and quality guarantees.
Our work can be viewed as a complementary direction to previous research trends in dense CRFs.
While~\cite{denseccp_kra,princip-parallel,vibhav-high-order} improved mean-field and~\cite{learn_cnn,crfasrnn} learnt the parameters, we focus on the energy minimisation problem.

\section{Preliminaries}
\label{sec:preli}
Before describing our methods for energy minimisation on dense CRF, we establish the necessary notation and background information.

\subsubsection{Dense CRF Energy Function.}\mbox{ }
We define a dense CRF on a set of $N$ random variables $\mathcal{X} = \left\{X_{1}, \dots, X_{N}\right\}$ each of which can take one label from a set of $M$ labels $\mathcal{L} = \left\{l_{1}, \dots\, l_{M}\right\}$.
To describe a labelling, we use a vector $\mathbf{x}$ of size $N$ such that its element $x_{a}$ is the label taken by the random variable $X_{a}$.
The energy associated with a given labelling is defined as:
\begin{equation}
\label{eq:labelling-energy}
E(\mathbf{x}) = \sum_{a=1}^{N}\phi_{a}(x_{a}) + \sum_{a=1}^{N}\sum_{\substack{b=1\\b \neq a}}^{N} \psi_{a,b}(x_{a}, x_{b}).
\end{equation}
Here, $\phi_{a}(x_{a})$ is called the \textit{unary potential} for the random variable $X_{a}$ taking the label $x_{a}$. The term $\psi_{a,b}(x_{a}, x_{b})$ is called the \textit{pairwise potential} for the random variables $X_{a}$ and $X_{b}$ taking the labels $x_{a}$ and $x_{b}$ respectively.
The energy minimisation problem on this CRF can be written as:
\begin{equation}
\label{eq:map-pb}
\mathbf{x}^{\star} = \argmin_{\mathbf{x}} E(\mathbf{x}).
\end{equation}

\subsubsection{Gaussian Pairwise Potentials.}\mbox{ }
Similar to previous work~\cite{densecrf-kra}, we consider arbitrary unary potentials and Gaussian pairwise potentials. Specifically, the form of the pairwise potentials is given by:
\begin{equation}
\label{eq:pairwise}
\psi_{a,b}(i,j) = \mu(i, j) \sum_{m} w^{(m)} k(\mathbf{f}^{(m)}_{a}, \mathbf{f}^{(m)}_{b}),
\end{equation}
\begin{equation}
  \label{eq:gaussiankernel}
  k(\mathbf{f}_{a}, \mathbf{f}_{b}) = \exp\left(\frac{-\|\mathbf{f}_{a} - \mathbf{f}_{b}\|^{2}}{2}\right)
\end{equation}
We refer to the term $\mu(i,j)$ as a \textit{label compatibility} function between the labels $i$ and $j$.
An example of a label compatibility function is the Potts model, where $\mu_{\text{potts}}(i, j) = [i \neq j]$, that is $\mu_{potts}(i,j) = 1$ if $i \neq j$ and 0 otherwise.
Note that the label compatibility does not depend on the image.
The other term, called the \textit{pixel compatibility} function, is a mixture of gaussian kernels $k(\cdot,\cdot)$.
The coefficients of the mixture are the weights $w^{(m)}$.
The $\mathbf{f}^{(m)}_{a}$ are the features describing the random variable $X_{a}$.
Note that the pixel compatibility does not depend on the labelling.
In practice, similar to~\cite{densecrf-kra}, we use the position and RGB values of a pixel as features.

\subsubsection{IP Formulation.}\mbox{ }
We now introduce a formulation of the energy minimisation problem that is more amenable to continuous relaxations.
Specifically, we formulate it as an Integer Program (IP) and then relax it to obtain a continuous optimisation problem.
To this end, we define the vector $\mathbf{y}$ whose components $y_{a}(i)$ are indicator variables specifying whether or not the random variable $X_{a}$ takes the label $i$.
Using this notation, we can rewrite the energy minimisation problem as an IP:
\begin{equation}
  \label{eq:energy-ip}
  \begin{split}
    \min \quad& \sum_{a=1}^{N} \sum_{i \in \mathcal{L}} \phi_{a}(i) y_{a}(i) + \sum_{a=1}^{N} \sum_{\substack{b=1\\b \neq a}}^{N} \sum_{i,j \in \mathcal{L}} \psi_{a,b}(i,j) y_{a}(i)y_{b}(j),\\
    \text{s.t. }& \sum_{i \in \mathcal{L}} y_{a}(i) = 1 \quad \forall a \in [1, N],\\
    & y_{a}(i) \in \{0,1\} \quad \forall a \in [1, N] \quad \forall i \in \mathcal{L}.\\
  \end{split}
\end{equation}
The first set of constraints model the fact that each random variable has to be assigned exactly one label.
The second set of constraints enforce the optimisation variables $y_{a}(i)$ to be binary.
Note that the objective function is equal to the energy of the labelling encoded by $\mathbf{y}$.

\subsubsection{Filter-based Method.}\mbox{ }
Similar to~\cite{densecrf-kra}, a key component of our algorithms is the filter-based method of Adams et al.~\cite{permuto}. It computes the following operation:
\begin{equation}
\label{eq:fb}
\forall a \in [1, N], \quad v^{\prime}_{a} = \sum_{b=1}^{N} k(\mathbf{f}_{a}, \mathbf{f}_{b}) v_{b},
\end{equation}
where $v_{a}^{\prime}, v_{b} \in \mathbb{R}$ and $k(\cdot,\cdot)$ is a Gaussian kernel.
Performing this operation the na\"{i}ve way would result in computing a sum on $N$ elements for each of the $N$ terms that we want to compute.
The resulting complexity would be $\mathcal{O}(N^2)$.
The filter-based method allows us to perform it approximately with $\mathcal{O}(N)$ complexity.
We refer the interested reader to~\cite{permuto} for details.
The accuracy of the approximation made by the filter-based method is explored in the Appendix~\ref{sec:supp-approx}.

\section{Quadratic Programming Relaxation}
\label{sec:QP}
We are now ready to demonstrate how the filter-based method~\cite{permuto} can be used to optimise our first continuous relaxation, namely the convex quadratic programming (QP) relaxation.

\subsubsection{Notation.}\mbox{ }
In order to concisely specify the QP relaxation, we require some additional notation.
Similar to ~\cite{denseccp_kra}, we rewrite the objective function with linear algebra operations.
The vector $\bm{\phi}$ contains the unary terms.
The matrix $\bm{\mu}$ corresponds to the label compatibility function.
The Gaussian kernels associated with the $m$-th features are represented by their Gram matrix \mbox{$\mathbf{K}^{(m)}_{a,b} = k(\mathbf{f}_{a}^{(m)}, \mathbf{f}_{b}^{(m)})$}.
The Kronecker product is denoted by $\otimes$. The matrix $\bm{\Psi}$ represents the pairwise terms and is defined as follows:
\begin{equation}
\label{eq:pairwise-matrix}
\bm{\Psi} = \bm{\mu} \otimes \left( \sum_{m} \mathbf{K}^{(m)} - \mathbf{I}_{N} \right),
\end{equation}
where $\mathbf{I}_{N}$ is the identity matrix.
Under this notation, the IP~(\ref{eq:energy-ip}) can be concisely written as
\begin{equation}
\label{eq:matrix-ilp}
\begin{split}
\min \quad&\bm{\phi}^{T} \mathbf{y} + \mathbf{y}^{T} \bm{\Psi} \mathbf{y},\\
\text{ s.t. } \quad&\mathbf{y} \in \mathcal{I},
\end{split}
\end{equation}
with $\mathcal{I}$ being the feasible set of integer solution, as defined in equation~(\ref{eq:energy-ip}).

\subsubsection{Relaxation.}\mbox{ }
In general, IP such as (\ref{eq:matrix-ilp}) are NP-hard problems. Relaxing the integer constraint on the indicator variables to allow fractional values between 0 and 1 results in the QP formulation.
Formally, the feasible set of our minimisation problem becomes:
\begin{equation}
\label{eq:feasible-set}
\mathcal{M} = \left\{\mathbf{y} \quad \text{such that } \quad \begin{aligned} \sum_{i \in \mathcal{L}} y_{a}(i) &= 1 \quad &&\forall a \in [1, N],\\
y_{a}(i) &\geq 0 &&\forall a \in [1, N], \forall i \in \mathcal{L} \end{aligned}\right\}.
\end{equation}
Ravikumar and Lafferty~\cite{ravikumar} showed that this relaxation is tight and that solving the QP will result in solving the IP.
However, this QP is still NP-hard, as the objective function is non-convex.
To alleviate this difficulty, Ravikumar and Lafferty~\cite{ravikumar} relaxed the QP minimisation to the following convex problem:
\begin{equation}
\label{eq:cvx-qp-min}
\begin{split}
\min \quad&S_{cvx}(\mathbf{y}) = (\bm{\phi} - \mathbf{d})^{T} \mathbf{y} + \mathbf{y}^{T}(\bm{\Psi} + \mathbf{D}) \mathbf{y},\\
\text{s.t. } \quad& \mathbf{y} \in \mathcal{M},
\end{split}
\end{equation}
where the vector $\mathbf{d}$ is defined as follows
\begin{equation}
\label{eq:RL-diagonal}
d_{a}(i) = \sum_{\substack{b=1 \\ b \neq a}}^{N} \sum_{j \in \mathcal{L}} |\psi_{a,b}(i,j)|,
\end{equation}
and $\mathbf{D}$ is the square diagonal matrix with $\mathbf{d}$ as its diagonal.

\subsubsection{Minimisation.}\mbox{ }
We now introduce a new method based on the Frank-Wolfe algorithm~\cite{frank-wolfe} to minimise problem (\ref{eq:cvx-qp-min}). The Frank-Wolfe algorithm allows to minimise a convex function $f$ over a convex feasible set $\mathcal{M}$. The key steps of the algorithm are shown in Algorithm \ref{alg:fw}.
To be able to use the Frank-Wolfe algorithm, we need a way to compute the gradient of the objective function (Step \ref{line:gradient}), a method to compute the conditional gradient (Step \ref{line:lp}) and a strategy to choose the step size (Step \ref{line:alpha}).

\begin{algorithm}
  \begin{algorithmic}[1]
    \State Get $\mathbf{y}^{0} \in \mathcal{M}$
    \While{not converged}
         \State Compute the gradient at $\mathbf{y}^{t}$ as $\mathbf{g} = \nabla f(\mathbf{y}^{t})$ \label{line:gradient}
         \State Compute the conditional gradient as $\mathbf{s} = \argmin_{\mathbf{s} \in \mathcal{M}} \langle \mathbf{s}, \mathbf{g} \rangle$ \label{line:lp}
         \State Compute a step-size $\alpha = \argmin_{\alpha \in [0,1]} f(\alpha \mathbf{y}^{t} + (1-\alpha) \mathbf{s}) $ \label{line:alpha}
         \State Move towards the negative conditional gradient $ \mathbf{y}^{t+1} = \alpha \mathbf{y}^{t} + (1 - \alpha) \mathbf{s} $ \label{line:update}
    \EndWhile
  \end{algorithmic}
  \caption{\label{alg:fw} Frank-Wolfe algorithm}
\end{algorithm}

\paragraph{Gradient computation}\mbox{} \\
Since the objective function is quadratic, its gradient can be computed as
\begin{equation}
\label{eq:LRQP-grad}
\nabla S_{\text{cvx}}(\mathbf{y}) = (\bm{\phi} - \mathbf{d}) + 2 (\bm{\Psi} + \mathbf{D}) \mathbf{y}.
\end{equation}
What makes this equation expensive to compute in a na\"{i}ve way is the matrix product with $\bm{\Psi}$.
We observe that this operation can be performed using the filter-based method in linear time.
Note that the other matrix-vector product, \mbox{$\mathbf{D} \mathbf{y}$}, is not  expensive (linear in N) since $\mathbf{D}$ is a diagonal matrix.

\paragraph{Conditional gradient}\mbox{} \\
The conditional gradient is obtained by solving
\begin{equation}
  \label{eq:cond-grad-pb}
\argmin_{\mathbf{s} \in \mathcal{M}} \langle \mathbf{s}, \nabla S_{\text{cvx}}(\mathbf{y}) \rangle.
\end{equation}
Minimising such an LP would usually be an expensive operation for problems of this dimension.
However, we remark that, once the gradient has been computed, exploiting the properties of our problem allows us to solve problem~(\ref{eq:cond-grad-pb}) in a time linear in the number of random variables ($N$) and labels ($M$).
Specifically, the following is an optimal solution to problem~(\ref{eq:cond-grad-pb}).
\begin{equation}
\label{eq:LRQP-lp}
\mathbf{s}_{a}(i) = \begin{cases}
1 \quad \text{if } i = \argmin_{i \in \mathcal{L}} \frac{\partial S_{\text{cvx}}}{\partial y_{a}(i)}\\
0 \quad \text{else.}
\end{cases}.
\end{equation}

\paragraph{Step size determination}\mbox{} \\
In the original Frank-Wolfe algorithm, the step size $\alpha$ is simply chosen using line search.
However we observe that, in our case, the optimal $\alpha$ can be computed by solving a second-order polynomial function of a single variable, which has a closed form solution that can be obtained efficiently.
This observation has been previously exploited in the context of Structural SVM~\cite{jaggi-lacoste}.
The derivations for this closed form solution can be found in Appendix~\ref{supp-fw-close}.
With careful reutilisation of computations, this step can be performed without additional filter-based method calls.
By choosing the optimal step size at each iteration, we reduce the number of iterations needed to reach convergence.

The above procedure converges to the global minimum of the convex relaxation and resorts to the filter-based method only once per iteration during the computation of the gradient and is therefore efficient.
However, this solution has no guarantees to be even a local minimum of the original QP relaxation. To alleviate this, we will now introduce a difference-of-convex (DC) relaxation.

\section{Difference of Convex Relaxation}
\subsection{DC relaxation: General case}
\label{subsec:DC_gen}
The objective function of a general DC program can be specified as
\begin{equation}
\label{eq:cccp-decomp}
S_{\text{CCCP}}(\mathbf{y}) = p(\mathbf{y}) - q(\mathbf{y}).
\end{equation}
One can obtain one of its local minima using the Concave-Convex Procedure (CCCP)~\cite{yuille2002concave}. The key steps of this algorithm are described in Algorithm \ref{alg:cccp}. Briefly, \mbox{Step \ref{line:cccp-lin}} computes the gradient of the concave part. \mbox{Step \ref{line:cccp-cvx}} minimises a convex upper bound on the DC objective, which is tight at  $\mathbf{y}^{t}$.

In order to exploit the CCCP algorithm for DC programs, we observe that the QP (\ref{eq:matrix-ilp}) can be rewritten as
\begin{equation}
\label{eq:cccp-qp}
\begin{split}
\min_{\mathbf{y}} \quad&\bm{\phi}^{T} \mathbf{y} + \mathbf{y}^{T} (\bm{\Psi + \mathbf{D}}) \mathbf{y} - \mathbf{y}^{T}\mathbf{D}\mathbf{y},\\
\text{ s.t. } \quad&\mathbf{y} \in \mathcal{M}.
\end{split}
\end{equation}
Formally, we can define \mbox{$p(\mathbf{y}) = \bm{\phi}^{T} \mathbf{y} + \mathbf{y}^{T}(\bm{\Psi} + \mathbf{D}) \mathbf{y}$} and \mbox{$q(\mathbf{y}) = \mathbf{y}^{T} \mathbf{D} \mathbf{y}$}, which are both convex in $\mathbf{y}$.

\begin{algorithm}
  \begin{algorithmic}[1]
    \State Get $\mathbf{y}^{0} \in \mathcal{M}$
    \While{not converged}
         \State Linearise the concave part $\mathbf{g} = \nabla q(\mathbf{y}^{t})$ \label{line:cccp-lin}
         \State Minimise a convex upper-bound $ \mathbf{y}^{t+1} = \argmin_{\mathbf{y} \in \mathcal{M}} p(\mathbf{y}) - \mathbf{g}^{T} \mathbf{y}$ \label{line:cccp-cvx}
    \EndWhile
  \end{algorithmic}
  \caption{\label{alg:cccp} CCCP Algorithm}
\end{algorithm}

We observe that, since $\mathbf{D}$ is diagonal and the matrix product with $\bm{\Psi}$ can be computed using the filter based method, the gradient $\nabla q(\mathbf{y}^{t}) = 2 \mathbf{D} \mathbf{y}$ (Step \ref{line:cccp-lin}) is efficient to compute.
The minimisation of the convex problem \mbox{(Step \ref{line:cccp-cvx})} is analogous to the convex QP formulation (\ref{eq:cvx-qp-min}) presented above with different unary potentials. Since we do not place any restrictions on the form of the unary potentials, \mbox{(Step \ref{line:cccp-cvx})} can be implemented using the method described in \mbox{Section \ref{sec:QP}}.

The CCCP algorithm provides a monotonous decrease in the objective function and will converge to a local minimum~\cite{DBLP:conf/nips/SriperumbudurL09}. However, the above method will take several iterations to converge, each necessitating the solution of a convex QP, and thus requiring multiple calls to the filter-based method. While the filter-based method \cite{permuto} allows us to compute operations on the pixel compatibility function in linear time, it still remains an expensive operation to perform. As we show next, if we introduce some additional restriction on our potentials, we can obtain a more efficient difference of convex decomposition.

\subsection{DC relaxation: negative semi-definite compatibility}
\label{subsec:DC_neg}
We now introduce a new DC relaxation of our objective function that takes advantage of the structure of the problem. Specifically, the convex problem to solve at each iteration does not depend on the filter-based method computations, which are the expensive steps in the previous method.
Following the example of Kr{\"a}henb{\"u}hl and Koltun~\cite{denseccp_kra}, we look at the specific case of negative semi-definite label compatibility function, such as the commonly used Potts model. Taking advantage of the specific form of our pairwise terms (\ref{eq:pairwise-matrix}), we can rewrite the problem as
\begin{equation}
\label{eq:QP-concave-CCCP}
S(\mathbf{y}) = \bm{\phi}^{T} \mathbf{y} - \mathbf{y}^{T}(\bm{\mu} \otimes \mathbf{I}_{N}) \mathbf{y}^{T}
+ \mathbf{y}^{T} (\bm{\mu} \otimes \sum_{m} \mathbf{K}^{(m)}) \mathbf{y}.
\end{equation}
The first two terms can be verified as being convex.
The Gaussian kernel is positive semi-definite, so the Gram matrices $\mathbf{K}^{(m)}$ are positive semi-definite.
By assumption, the label compatibility function is also negative semi-definite.
The results from the Kronecker product between the Gram matrix and $\mathbf{\mu}$ is therefore negative semi-definite.

\subsubsection{Minimisation.}\mbox{ }
Once again we use the CCCP Algorithm. The main difference between the generic DC relaxation and this specific one is that \mbox{Step \ref{line:cccp-lin}} now requires a call to the filter-based method, while the iterations required to solve \mbox{Step \ref{line:cccp-cvx}} do not. In other words, each iteration of CCCP only requires one call to the filter based method. This results in a significant improvement in speed.
More details about this operation are available in Appendix~\ref{supp-dcneg}.

\section{LP relaxation}
\label{sec:LP}
This section presents an accurate LP relaxation of the energy minimisation problem and our method to optimise it efficiently using subgradient descent.
\subsubsection{Relaxation.}\mbox{ }
To simplify the description, we focus on the Potts model.
However, our approach can easily be extended to more general pairwise potentials by approximating them using a hierarchical Potts model.
Such an extension, inspired by~\cite{kumar2009map}, is presented in Appendix~\ref{supp-genpairwise}.
We define the following notation: $K_{a,b} = \sum_{m} w^{(m)}k^{(m)}(\mathbf{f}^{(m)}_{a}$, $\mathbf{f}^{(m)}_{b})$, $\sum_{a} = \sum_{a=1}^{N}$ and $\sum_{b<a} = \sum_{b=1}^{a-1}$.
With these notations, a LP relaxation of (\ref{eq:energy-ip}) is:
\begin{equation}
\label{eq:potts-lp}
    \begin{split}
        \min \quad& S_{LP}(\mathbf{y}) = \underbrace{\sum_{a} \sum_{i} \phi_{a}(i) y_{a}(i)}_{unary} + \underbrace{\sum_{a} \sum_{b \neq a} \sum_{i} K_{a,b} \frac{|y_{a}(i) - y_{b}(i)|}{2}}_{pairwise}, \\
        \text{s.t.} \quad& \mathbf{y} \in \mathcal{M}. \\
    \end{split}
\end{equation}
The feasible set remains the same as the one we had for the QP and DC relaxations.
In the case of integer solutions, $S_{LP}(\mathbf{y})$ has the same value as the objective function of the IP described in~(\ref{eq:energy-ip}).
The $unary$ term is the same for both formulations.
The $pairwise$ term ensures that for every pair of random variables \mbox{$X_{a}, X_{b}$}, we add the cost $K_{a,b}$ associated with this edge only if they are not associated with the same labels.

\subsubsection{Minimisation.}\mbox{ }
Kleinberg and Tardos~\cite{kleinbergTardos2002} solve this problem by introducing extra variables for each pair of pixels to get a standard LP, with a linear objective function and linear constraints.
In the case of a dense CRF, this is infeasible because it would introduce a number of variables quadratic in the number of pixels.
We will instead use projected subgradient descent to solve this LP.
To do so, we will reformulate the objective function, derive the subgradient, and present an algorithm to compute it efficiently.

\paragraph{Reformulation}\mbox{} \\
The absolute value in the pairwise term of~(\ref{eq:energy-ip}) prevents us from using the filtering approach.
To address this issue, we consider that for any given label $i$, the variables $y_{a}(i)$ can be sorted in a descending order: \mbox{$a \geq b \implies y_{a}(i) \leq y_{b}(i)$}.
This allows us to rewrite the pairwise term of the objective function (\ref{eq:potts-lp}) as:
\begin{equation}
  \label{eq:simple_S}
    \sum_{i} \sum_{a} \sum_{a \neq b} K_{a,b} \frac{|y_{a}(i) - y_{b}(i)|}{2}
   = \sum_{i} \sum_{a} \sum_{b > a} K_{a,b} y_{a}(i) - \sum_{i} \sum_{a} \sum_{b < a} K_{a,b} y_{a}(i).
\end{equation}
A formal derivation of this equality can be found in Appendix~\ref{supp-lp-reform}.

\paragraph{Subgradient}\mbox{} \\
From (\ref{eq:simple_S}), we rewrite the subgradient:
\begin{equation}
    \label{eq:grad_lp}
    \frac{\partial S_{LP}}{\partial y_{c}(k)}(\mathbf{y}) = \phi_{c}(k) + \sum_{a > c} K_{a,c} - \sum_{a < c} K_{a,c} . \\
\end{equation}
Note that in this expression, the dependency on the variable $\mathbf{y}$ is hidden in the bounds of the sum because we assumed that $y_{a}(k) \leq y_{c}(k)$ for all $a>c$.
For a different value of $\mathbf{y}$, the elements of $\mathbf{y}$ would induce a different ordering and the terms involved in each summation would not be the same.

\paragraph{Subgradient computation}\mbox{} \\
What prevents us from evaluating~(\ref{eq:grad_lp}) efficiently are the two sums, one over an upper triangular matrix ($\sum_{a > c} K_{a,c}$) and one over a lower triangular matrix ($\sum_{a<c} K_{a,c}$).
As opposed to~(\ref{eq:fb}), which computes terms $\sum_{a,b}K_{a,b}v_b$ for all $a$ using the filter-based method, the summation bounds here depend on the random variable we are computing the partial derivative for.
While it would seems that the added sparsity provided by the upper and lower triangular matrices would simplify the operation, it is this sparsity itself that prevents us from interpreting the summations as convolution operations.
Thus, we cannot use the filter-based method as described by Adams et al.~\cite{permuto}.

We alleviate this difficulty by designing a novel divide-and-conquer algorithm.
We describe our algorithm for the case of the upper triangular matrix.
However, it can easily be adapted to compute the summation corresponding to the lower triangular matrix.
We present the intuition behind the algorithm using an example.
A rigorous development can be found in Appendix~\ref{supp-lp-dc}.
If we consider $N=6$ then $a,c \in \{1,2,3,4,5,6\}$ and the terms we need to compute for a given label are:
\begin{equation}
  \scriptsize
\renewcommand\arraystretch{1.5}
\left(
  \begin{array}{c}
    \sum_{a > 1} K_{a,1}\\
    \sum_{a > 2} K_{a,2}\\
    \sum_{a > 3} K_{a,3}\\
    \sum_{a > 4} K_{a,4}\\
    \sum_{a > 5} K_{a,5}\\
    \sum_{a > 6} K_{a,6}\\
  \end{array}
\right)
=
\underbrace{
\left(
  \begin{array}{ccc|ccc}
    \enskip\text{ }0\text{ }\enskip & K_{2,1} & K_{3,1} & K_{4,1} & K_{5,1} & K_{6,1}\\
    0 & 0 & K_{3,2} & K_{4,2} & K_{5,2} & K_{6,2}\\
    0 & 0 & 0 & K_{4,3} & K_{5,3} & K_{6,3}\\
    \hline
    0 & 0 & 0 & 0 & K_{5,4} & K_{6,4}\\
    0 & 0 & 0 & 0 & 0 & K_{6,5}\\
    0 & 0 & 0 & 0 & 0 & 0\\
  \end{array}
\right)
}_{\mathbf{U}}
\cdot
\left(
  \begin{array}{c}
    1 \\
    1 \\
    1 \\
    1 \\
    1 \\
    1 \\
  \end{array}
\right)
\normalsize
\end{equation}
We propose a divide and conquer approach that solves this problem by splitting the upper triangular matrix $\mathbf{U}$.
The top-left and bottom-right parts are upper triangular matrices with half the size.
We solve these subproblems recursively.
The top-right part can be computed with the original filter based method.
Using this approach, the total complexity to compute this sum is $\mathcal{O}(N \log( N ))$.

With this algorithm, we have made feasible the computation of the subgradient.
We can therefore perform projected subgradient descent on the LP objective efficiently.
Since we need to compute the subgradient for each label separately due to the necessity of having sorted elements, the complexity associated with taking a gradient step is $\mathcal{O}(M N \log( N ))$.
To ensure the convergence, we choose as learning rate $(\beta^{t})_{t=1}^{\infty}$ that is a square summable but not a summable sequence such as $(\frac{1}{1+t})_{t=1}^{\infty}$.
We also make use of the work by Condat \cite{Condat2015} to perform fast projection on the feasible set.
The complete procedure can be found in Algorithm \ref{alg:lp_subgrad}.
Step~\ref{line:label_loop} to~\ref{line:label_loop_end} present the subgradient computation for each label.
Using this subgradient, Step~\ref{line:gradient_step} shows the update rule for $y^{t}$.
Finally, Step~\ref{line:proj} project this new estimate onto the feasible space.
\begin{algorithm}
  \begin{algorithmic}[1]
    \State Get $\mathbf{y}^{0} \in \mathcal{M}$
    \While{not converged}
      \For{$i \in \mathcal{L}$} \label{line:label_loop}
        \State Sort $y_{a}(i) \quad \forall a \in [1, N]$
        \State Reorder $\mathbf{K}$
        \State $\mathbf{g}(i) = \nabla S_{LP}(\mathbf{y}^{t}(i))$
    \EndFor \label{line:label_loop_end}
    \State $\mathbf{y}^{t+1} = \mathbf{y}^{t} - \beta^{t} \cdot \mathbf{g}$ \label{line:gradient_step}
    \State Project $\mathbf{y}^{t+1}$ on the feasible space \label{line:proj}
    \EndWhile
  \end{algorithmic}
  \caption{\label{alg:lp_subgrad} LP subgradient descent}
\end{algorithm}

The algorithm that we introduced converges to a global minimum of the LP relaxation.
By using the rounding procedure introduced by Kleinberg and Tardos \cite{kleinbergTardos2002}, it has a multiplicative bound of 2 for the dense CRF labelling problem on Potts models and $\mathcal{O}(\log( M ))$ for metric pairwise potentials.

\FloatBarrier
\section{Experiments}
\label{sec:xp}
We now demonstrate the benefits of using continuous relaxations of the energy minimisation problem on two applications: stereo matching and semantic segmentation.
We provide results for the following methods: the Convex QP relaxation ($\mathbf{QP_{cvx}}$), the generic and negative semi-definite specific DC relaxations ($\mathbf{DC_{gen}}$ and $\mathbf{DC_{neg}}$) and the LP relaxation (\textbf{LP}).
We compare solutions obtained by our methods with the mean-field baseline (\textbf{MF}).

\subsection{Stereo matching}
\subsubsection{Data. }
We compare these methods on images extracted from the Middlebury stereo matching dataset~\cite{middlebury_stereo}. The unary terms are obtained using the absolute difference matching function of~\cite{middlebury_stereo}.
The pixel compatibility function is similar to the one used by Kr\"{a}henb\"{u}hl and Koltun~\cite{densecrf-kra} and is described in Appendix~\ref{supp-compat-function}.
The label compatibility function is a Potts model.
\subsubsection{Results. }
\begin{figure}
  \begin{subfigure}{0.6\textwidth}
    \includegraphics[width=\columnwidth]{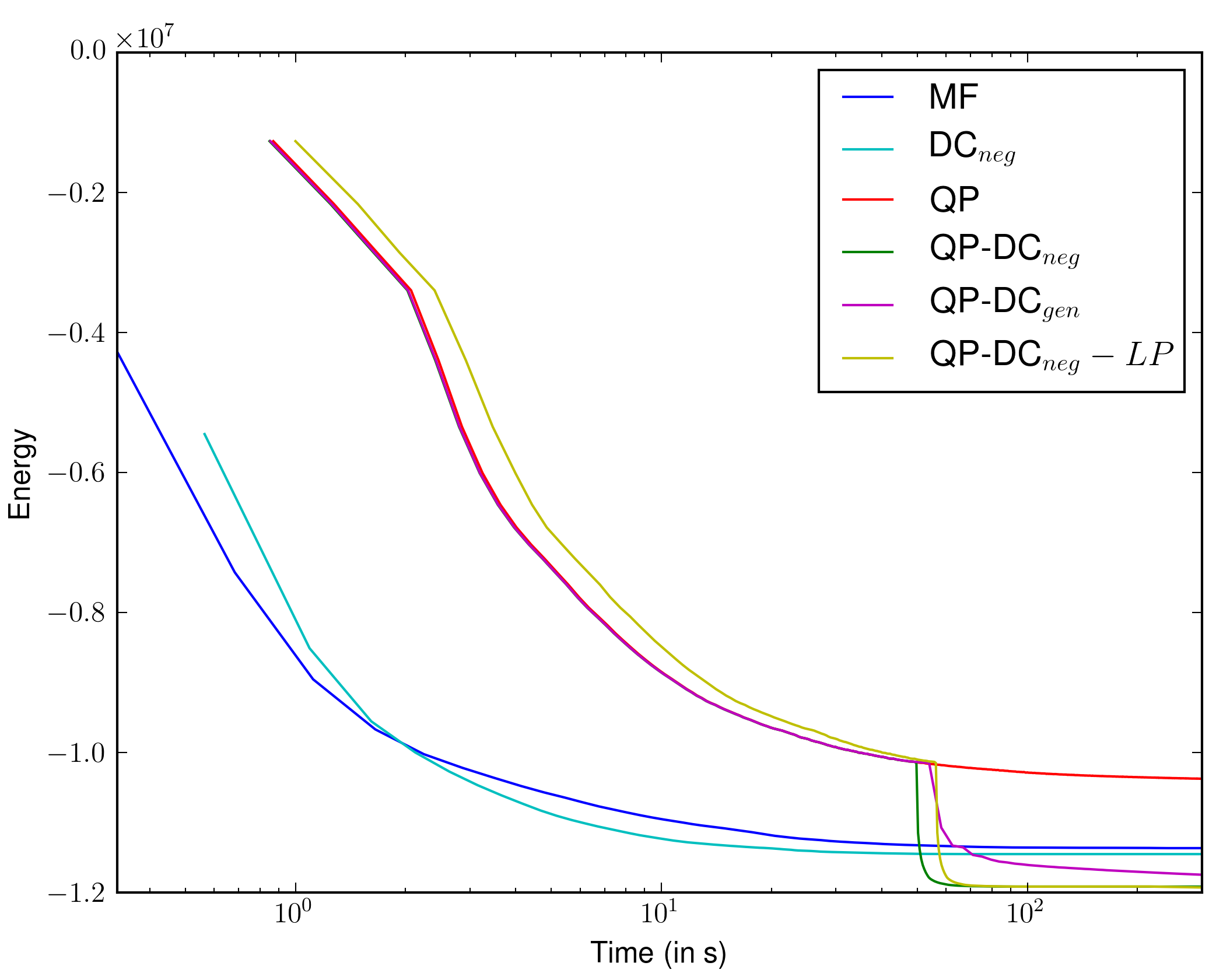}
    \caption{Runtime comparisons}
    \label{fig:runtime-comp}
  \end{subfigure}%
  \begin{subtable}{0.4\textwidth}
    \centering
  \begin{tabular}{|c|c|}
    \hline
    \textbf{Method} & \textbf{Final energy}\\
    \hline
    MF & -1.137e+07\\
    DC$_{neg}$ & -1.145e+07\\
    QP & -1.037e+07 \\
    QP-DC$_{neg}$ & -1.191e+07 \\
    QP-DC$_{gen}$ & -1.175e+07 \\
    QP-DC$_{neg}$-LP & \textbf{-1.193e+07}\\
    \hline
  \end{tabular}
\caption{Final Energy achieved}
  \label{tab:mid_nrj}
\end{subtable}
\caption{\em{Evolution of achieved energies as a function of time on a stereo matching problem (Teddy Image). While the \textbf{QP} method leads to the worst result, using it as an initialisation greatly improves results. In the case of negative semi-definite potentials, the specific \textbf{DC$_{neg}$} method is as fast as mean-field, while additionally providing guarantees of monotonous decrease.(Best viewed in colour}}
\end{figure}

We present a comparison of runtime in Figure (\ref{fig:runtime-comp}), as well as the associated final energies for each method in Table~(\ref{tab:mid_nrj}). Similar results for other problem instances can be found in Appendix~\ref{supp-stereo-res}.

We observe that continuous relaxations obtain better energies than their mean-field counterparts. For a very limited time-budget, \textbf{MF} is the fastest method, although \textbf{DC$_{neg}$} is competitive and reach lower energies.
When using \textbf{LP}, optimising a better objective function allows us to escape the local minima to which $\mathbf{DC_{neg}}$ converges.
However, due to the higher complexity and the fact that we need to perform divide-and-conquer separately for all labels, the method is slower.
This is particularly visible for problems with a high number of labels. This indicates that the LP relaxation might be better suited to fine-tune accurate solutions obtained by faster alternatives.
For example, this can be achieved by restricting the LP to optimise over a subset of relevant labels, that is, labels that are present in the solutions provided by other methods.
Qualitative results for the Teddy image can be found in Figure~\ref{fig:ted_stereo} and additional outputs are present in Appendix~\ref{supp-stereo-res}. We can see that lower energy translates to better visual results: note the removal of the artifacts in otherwise smooth regions (for example, in the middle of the sloped surface on the left of the image).

\begin{figure}[]
\captionsetup{font=footnotesize}
  \centering
\scalebox{0.9}{%
  \begin{tabular}{cccc}
  Left Image & \textbf{MF} & $\mathbf{DC_{neg}}$ & $\mathbf{QP_{cvx}}$ \\
  \includegraphics[width=0.2\columnwidth]{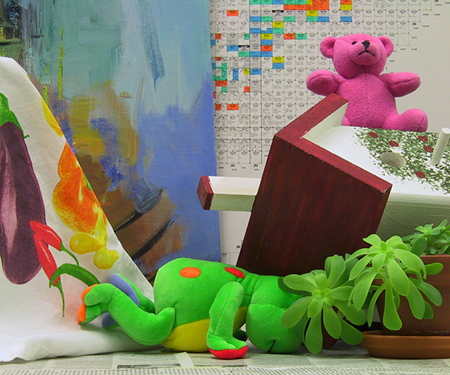} &
  \includegraphics[width=0.2\columnwidth]{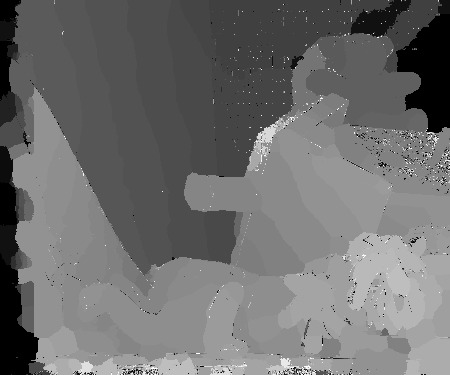} &
  \includegraphics[width=0.2\columnwidth]{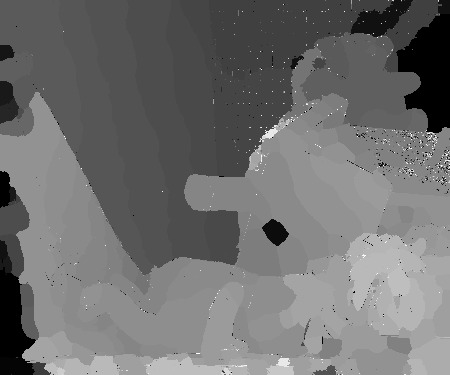} &
  \includegraphics[width=0.2\columnwidth]{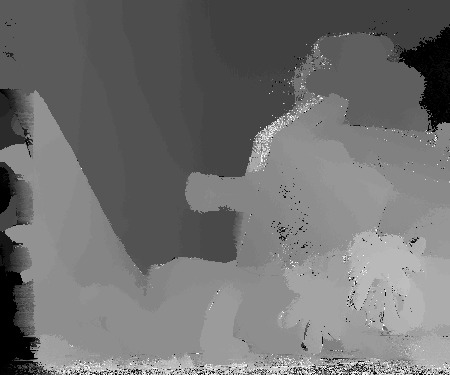} \\
  \includegraphics[width=0.2\columnwidth]{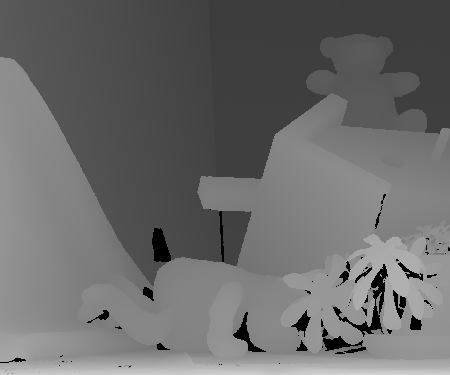} &
  \includegraphics[width=0.2\columnwidth]{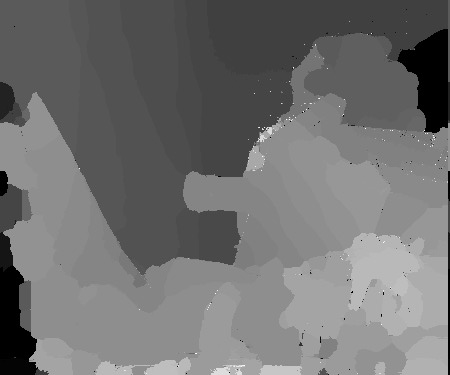} &
  \includegraphics[width=0.2\columnwidth]{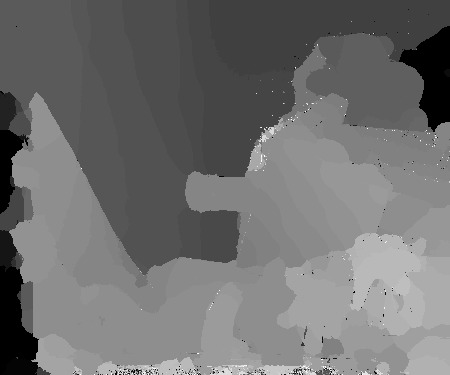} &
  \includegraphics[width=0.2\columnwidth]{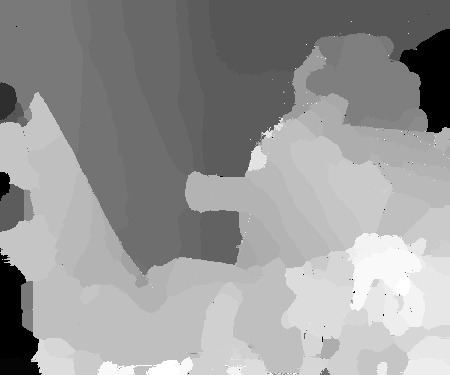} \\
  Ground Truth & $\mathbf{QP-DC_{neg}}$ & $\mathbf{QP-DC_{gen}}$ & $\mathbf{QP-DC_{neg}-LP}$ \\
  \end{tabular}
}
  \caption{Stereo matching results on the Teddy image. Continuous relaxation achieve smoother labeling, as expected by their lower energies.}
  \label{fig:ted_stereo}
\end{figure}

\subsection{Image Segmentation}
\subsubsection{Data. }
We now consider an image segmentation task evaluated on the PASCAL VOC 2010~\cite{pascal-voc-2010} dataset.
For the sake of comparison, we use the same data splits and unary potentials as the one used by Kr\"{a}henb\"{u}hl and Koltun~\cite{densecrf-kra}.
We perform cross-validation to select the best parameters of the pixel compatibility function for each method using Spearmint~\cite{snoek2012practical}.
\subsubsection{Results. }
The energy results obtained using the parameters cross validated for $\mathbf{DC_{neg}}$ are given in Table \ref{tab:cccpcv_res}.
\textbf{MF5} corresponds to mean-field ran for 5 iterations as it is often the case in practice~\cite{densecrf-kra,vibhav-high-order}.

\begin{table}[h]
\captionsetup{font=footnotesize}
\renewcommand{\arraystretch}{1.1}
\setlength{\tabcolsep}{0.7ex}
\centering
\begin{tabular}{@{}c@{\hspace*{2ex}}ccccccc@{\hspace*{3ex}}c@{\hspace*{3ex}}cc@{}}
\toprule
 & Unary & \textbf{MF5} & \textbf{MF} & $\mathbf{QP_{cvx}}$ & $\mathbf{DC_{gen}}$ & $\mathbf{DC_{neg}}$ & \textbf{LP} & \textbf{Avg. E} & \textbf{Acc} & \textbf{IoU}\\
\midrule
Unary & - & 0 & 0 & 0 & 0 & 0 & 0 & 0 & 79.04 & 27.43\\
\cmidrule(r{2ex}){2-8}\cmidrule(r{2ex}){9-9}\cmidrule{10-11}
\textbf{MF5} & 99 & - & 13 & 0 & 0 & 0 & 0 & -600 & 79.13 & 27.53\\
\textbf{MF} & 99 & 0 & - & 0 & 0 & 0 & 0 & -600 & 79.13 & 27.53\\
\cmidrule(r{2ex}){2-8}\cmidrule(r{2ex}){9-9}\cmidrule{10-11}
$\mathbf{QP_{cvx}}$ & 99 & 99 & 99 & - & 0 & 0 & 0 & -6014 & 80.38 & 28.56\\
$\mathbf{DC_{gen}}$ & 99 & 99 & 99 & 85 & - & 0 & 1 & -6429 & 80.41 & 28.59 \\
$\mathbf{DC_{neg}}$ & 99 & 99 & 99 & 98 & 97 & - & 4 & -6613 & 80.43 & 28.60 \\
\textbf{LP} & 99 & 99 & 99 & 98 & 97 & 87 & - & \textbf{-6697} & \textbf{80.49} & \textbf{28.68}\\
\bottomrule
\end{tabular}
\caption{Percentage of images the row method outperforms the column method on final energy, average energy over the test set and Segmentation performance. Continuous relaxations dominate mean-field approaches on almost all images and improve significantly more compared to the Unary baseline. Parameters tuned for $\mathbf{DC_{neg}}$. }
\label{tab:cccpcv_res}
\end{table}
Once again, we observe that continuous relaxations provide lower energies than mean-field based approaches. To add significance to this result, we also compare energies image-wise. In all but a few cases, the energies obtained by the continuous relaxations are better or equal to the mean-field ones.
This provides conclusive evidence for our central hypothesis that continuous relaxations are better suited to the problem of energy minimisation in dense CRFs.

For completeness, we also provide energy and segmentation results for the parameters tuned for \textbf{MF} in Appendix~\ref{supp-seg-nrj}.
Even in that unfavourable setting, continuous relaxations still provide better energies.
Note that, due to time constraints, we run the LP subgradient descent for only 5 iterations of subgradient descent.
Moreover, to be able to run more experiments, we also restricted the number of labels by discarding labels that have a very small probability to appear given the initialisation.

\begin{figure}[h]
\fontsize{3}{5}\selectfont
\captionsetup{font=footnotesize}
  \centering
  \begin{tabular}{@{}cc@{\hspace*{5ex}}cc@{\hspace*{5ex}}cc@{}}
  \multicolumn{2}{c}{Original} & \multicolumn{2}{c}{\textbf{MF} Parameters} & \multicolumn{2}{c}{$\mathbf{DC_{neg}}$ Parameters} \\[1ex]
  Image & Ground Truth & \textbf{MF} & $\mathbf{DC_{neg}}$ & \textbf{MF} & $\mathbf{DC_{neg}}$ \\[2ex]
  \includegraphics[width=0.14\columnwidth]{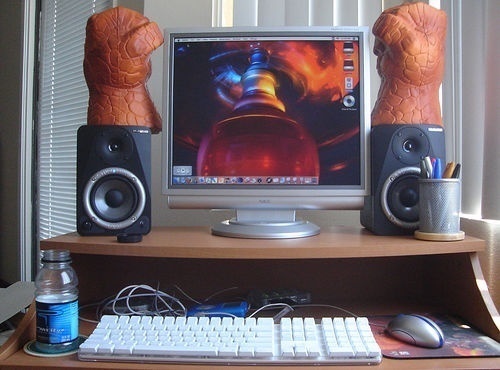} &
  \includegraphics[width=0.14\columnwidth]{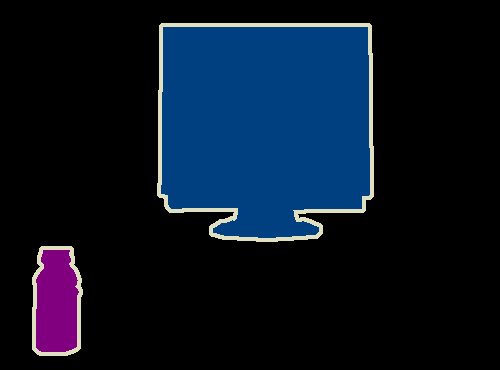} &
  \includegraphics[width=0.14\columnwidth]{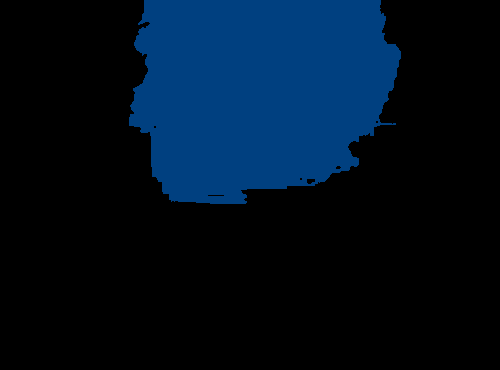} &
  \includegraphics[width=0.14\columnwidth]{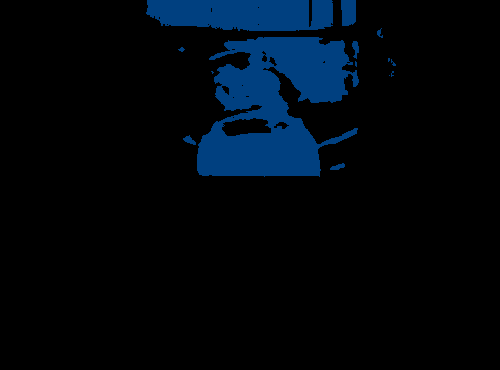} &
  \includegraphics[width=0.14\columnwidth]{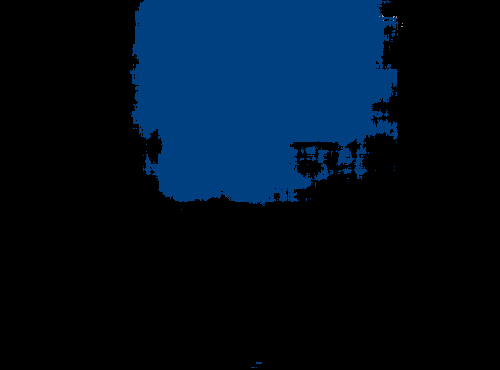} &
  \includegraphics[width=0.14\columnwidth]{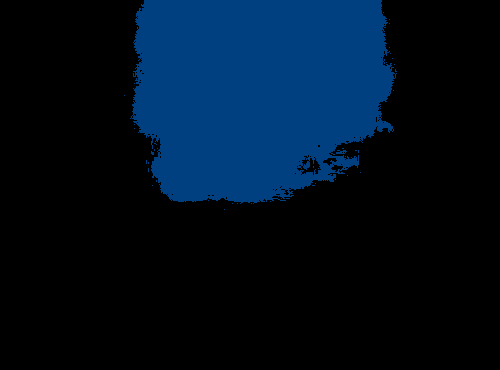} \\
  & & E=-3.08367e+7 & E=-3.1012e+7 & E=155992 & E=154100 \\[2ex]
  \includegraphics[width=0.14\columnwidth]{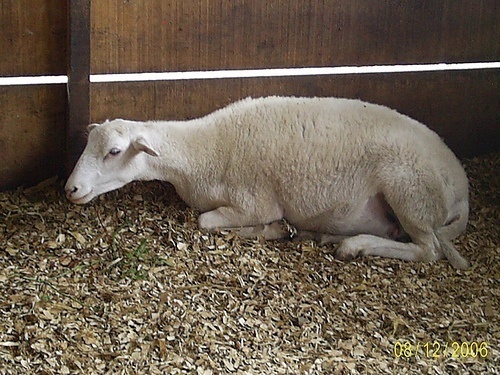} &
  \includegraphics[width=0.14\columnwidth]{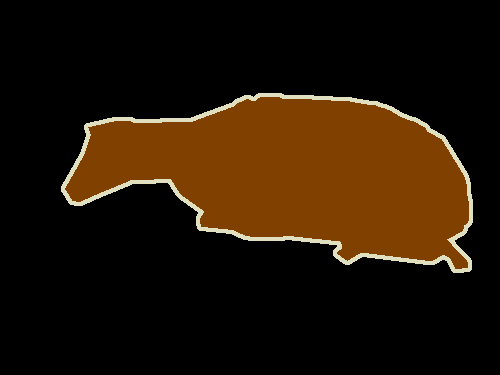} &
  \includegraphics[width=0.14\columnwidth]{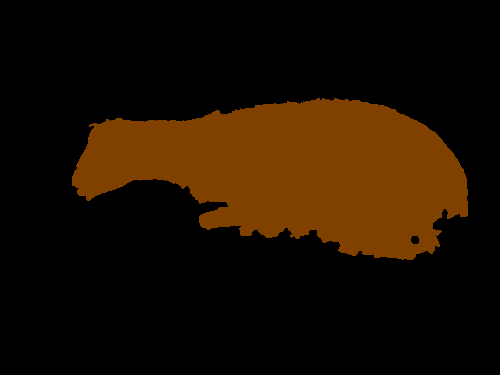} &
  \includegraphics[width=0.14\columnwidth]{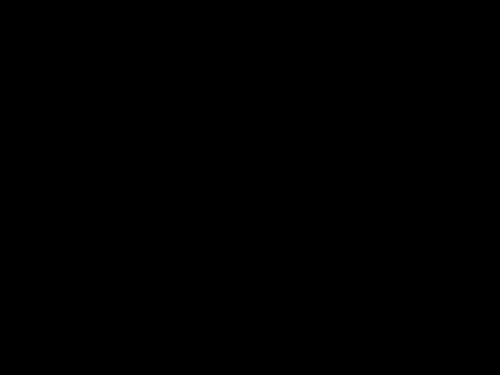} &
  \includegraphics[width=0.14\columnwidth]{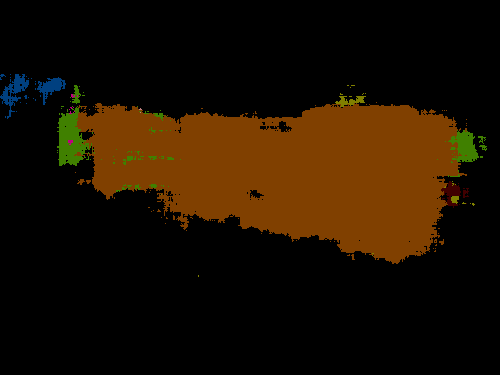} &
  \includegraphics[width=0.14\columnwidth]{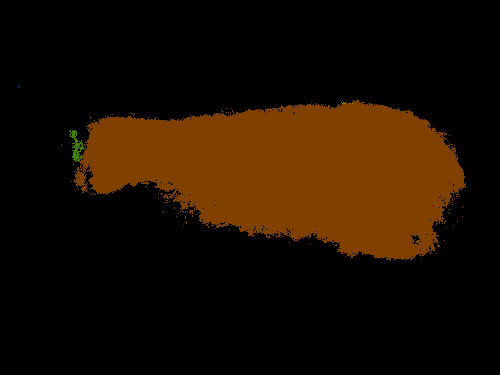} \\
  & & E=-3.10922e+7 & E=-3.18649e+7 & E=170968 & E=163308\\
  \end{tabular}
\normalsize
  \caption{Segmentation results on sample images. We see that $\mathbf{DC_{neg}}$ leads to better energy in all cases compared to \textbf{MF}. Segmentation results are better for \textbf{MF} for the \textbf{MF}-tuned parameters and better for $\mathbf{DC_{neg}}$ for the $\mathbf{DC_{neg}}$-tuned parameters.}
  \label{fig:seg_comp}
\end{figure}

Some qualitative results can be found in Figure~\ref{fig:seg_comp}.
When comparing the segmentations for \textbf{MF} and $\mathbf{DC_{neg}}$, we can see that the best one is always the one we tune parameters for.
A further interesting caveat is that although we always find a solution with better energy, it does not appear to be reflected in the quality of the segmentation.
While in the previous case with stereo vision, better energy implied qualitatively better reconstruction it is not so here.
Similar observation was made by Wang et al~\cite{Wang_2015_CVPR}.

\FloatBarrier
\section{Discussion}
Our main contribution are four efficient algorithms for the dense CRF energy minimisation problem based on QP, DC and LP relaxations.
We showed that continuous relaxations give better energies than the mean-field based approaches.
Our best performing method, the LP relaxation, suffers from its high runtime.
To go beyond this limit, move making algorithms such as $\alpha$-expansion~\cite{alpha_exp} could be used and take advantage of the fact that this relaxation solves exactly the original IP for the two label problem.
In future work, we also want to investigate the effect of learning specific parameters for these new inference methods using the framework of~\cite{crfasrnn}.

\footnotesize
\section*{Acknowledgments}
This work was supported by the EPSRC, Leverhulme Trust, Clarendon Fund and the ERC grant ERC-2012-AdG 321162-HELIOS, EPSRC/MURI grant ref EP/N019474/1, EPSRC grant EP/M013774/1, EPSRC Programme Grant Seebibyte EP/M013774/1 and Microsoft Research PhD Scolarship Program. We thank Philip Kr\"{a}henb\"{u}hl for making his code available and Vibhav Vineet for his help.
\normalsize

\bibliographystyle{splncs}
\bibliography{bibliography}

\section{Appendix}
\appendix
\label{sec:sup}
\section{Filter-based method approximation}
\label{sec:supp-approx}
In this paper, the filter based method that we use for our experiments is the one by Adams et al.~\cite{permuto}.
In this method, the original computation is approximated by a convolution in a higher dimensional space.
The original points are associated to a set of vertices on which the convolution is performed.
The considered vertices are the one from the permutohedral lattice.
Kr\"{a}henb\"{u}hl and Koltun~\cite{densecrf-kra} provided an implementation of this method.
In their implementation, they added a pixel-wise normalisation of the output of the permutohedral lattice and say that it performs well in practice.

\begin{figure}[]
\captionsetup{font=scriptsize}
  \centering
\scalebox{1}{%
  \begin{tabular}{cc}
  Full computation & Half computations \\
  \includegraphics[width=0.48\columnwidth]{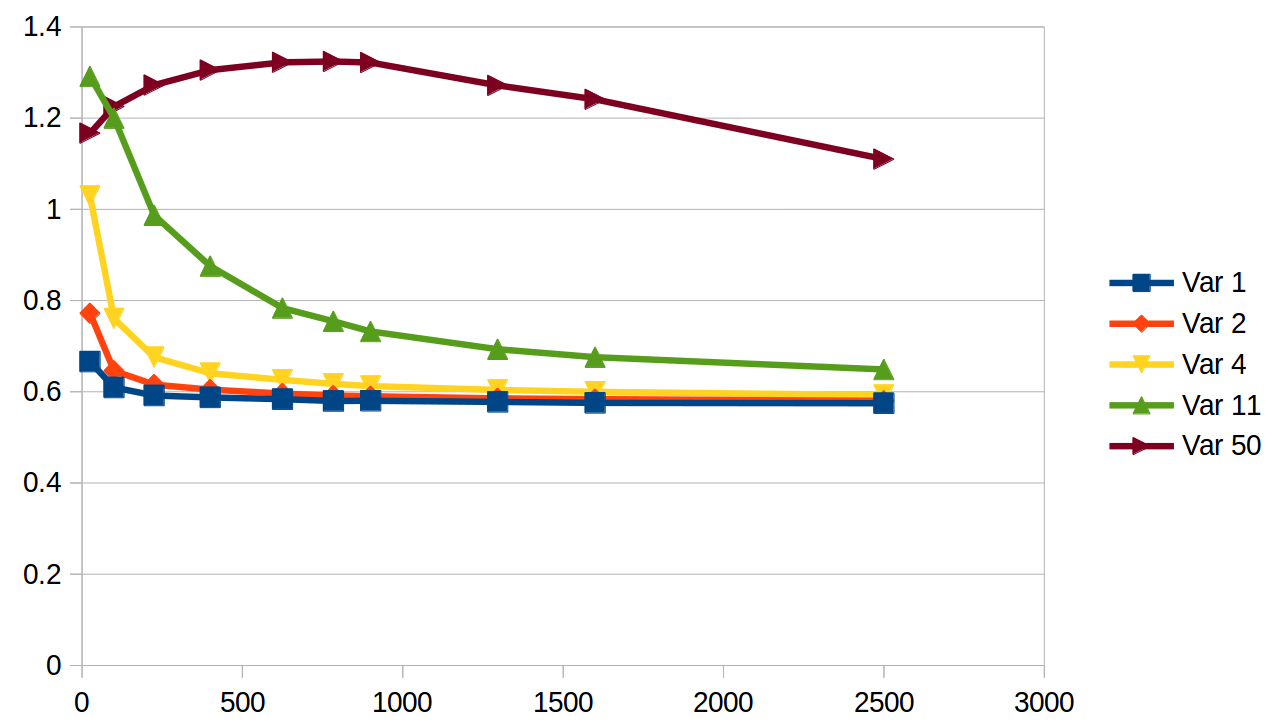} \quad&\quad
  \includegraphics[width=0.48\columnwidth]{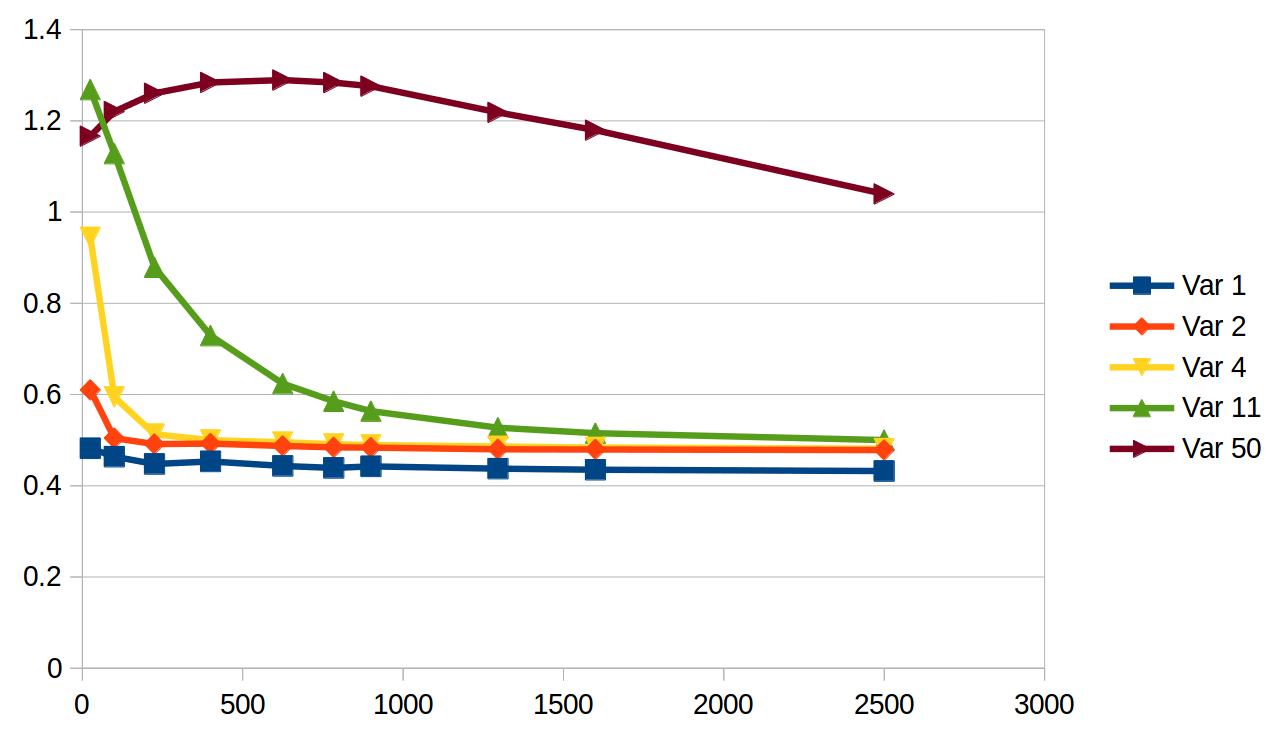} \\
  \end{tabular}
}
  \caption{Permutohedral lattice approximation. The vertical axis is the value computed in a brute force manner over the value computed by the permutohedral lattice. The horizontal axis is the number of pixels in the considered images. The graph on the left shows this ratio when comparing full permutohedral lattice computations with only ones as input. The graph on the right shows this ratio for the divide and conquer approach presented in the LP section. We can see that in both cases, for sufficiently large images, the permutohedral lattice computation is a constant factor away from the brute force value. This constant being a function of the input points values.}
  \label{fig:permu_plots}
\end{figure}

We observe that for the variances considered in this paper and \textbf{without} using the normalisation by Kr\"{a}henb\"{u}hl and Koltun, the results given by the permutohedral lattice is a constant factor away from the value computed by brute force in most cases.
As can be seen in Figure~\ref{fig:permu_plots}, in the case where we compute $\sum_{a,b} K_{a,b} 1$, the left graph, the ratio between the value obtained by brute force and the value obtained using the permutohedral lattice is $0.6$ for large enough images.
On the other hand, for a different value of the input points where we compute $\sum_{b>a} K_{a,b} - \sum_{b<a} K_{a,b}$, the right graph, we get a ratio of $0.48$ between the two results.
The case where we consider a variance of $50$ is special.
We know that the highest the variance value, the worst the approximation of the permutohedral is.
If the experience on the full computation is conducted on an image of size $320 \times 213$, the ratio between the brute force approach and the permutohedral lattice is $0.633$.
At the same time is also worth noting that in all these results, if we consider the outputs as vectors, as is done when computing our gradients, the vectors given by the brute force and the ones given by the permutohedral lattice are collinear for all image size and all variances.
We can thus expect that for other input values, the direction of gradient provided by the permutohedral lattice is correct, but the norm of this vector may be incorrect.

\section{Optimal step size in the Frank-Wolfe algorithm}
\label{supp-fw-close}
Solving the convex relaxation of the QP is performed using the Franke-Wolfe algorithm~\cite{frank-wolfe}.
The gradient is computed efficiently using the filter-based method~\cite{permuto}.
An efficient method is available to compute the conditional gradient, based on the gradient.
Once this conditional gradient is obtained, a step size needs to be determined to updates the value of the current parameters.
We show that the optimal step size can be computed and that this does not introduce any additional call to the filter-based method.

The problem to solve is
\begin{equation}
\argmin_{\alpha \in [0,1]} S_{cvx}(\mathbf{y} + \alpha (\mathbf{s} - \mathbf{y})).
\end{equation}
The definition of $S_{cvx}$ is
\begin{equation}
\label{eq:cvx-qp-min}
S_{cvx}(\mathbf{y}) = (\bm{\phi} - \mathbf{d})^{T} \mathbf{y} + \mathbf{y}^{T}(\bm{\Psi} + \mathbf{D}) \mathbf{y}.
\end{equation}
Solving for the optimal value of $\alpha$ amounts to solving a second order polynomial:
\begin{equation}
\label{eq:second-order-polynomial}
\begin{split}
S_{cvx}(\mathbf{y} + \alpha (\mathbf{s} - \mathbf{y}) &=
 (\bm{\phi} - \mathbf{d})^{T} (\mathbf{y} + \alpha (\mathbf{s} - \mathbf{y}))\\
& \qquad + (\mathbf{y} + \alpha (\mathbf{s} - \mathbf{y}))^{T}(\bm{\Psi} + \mathbf{D})  (\mathbf{y} + \alpha (\mathbf{s} - \mathbf{y})),\\
&= \alpha^{2} \left[(\mathbf{s}-\mathbf{y})^{T}(\bm{\Psi} + \mathbf{D})  (\mathbf{s}-\mathbf{y}) \right]\\
& \qquad + \alpha  \left[  (\bm{\phi} - \mathbf{d})^{T} (\mathbf{s} - \mathbf{y}) + 2 \mathbf{y}^{T} (\bm{\Psi} + \mathbf{D})(\mathbf{s} - \mathbf{y})  \right]\\
& \qquad + \quad \left[  (\bm{\phi} - \mathbf{d})^{T}\mathbf{y} + \mathbf{y}^{T} (\bm{\Psi} + \mathbf{D}) \mathbf{y}\right],
\end{split}
\end{equation}
whose optimal value is given by
\begin{equation}
\label{eq:alpha-opt}
\alpha^{\star} = -\frac{1}{2}  \frac{(\bm{\phi} - \mathbf{d})^{T} (\mathbf{s} - \mathbf{y}) + 2 \mathbf{y}^{T} (\bm{\Psi} + \mathbf{D})(\mathbf{s} - \mathbf{y})}{(\mathbf{s}-\mathbf{y})^{T}(\bm{\Psi} + \mathbf{D})  (\mathbf{s}-\mathbf{y})}
\end{equation}

The dot products are going to be linear in complexity and efficient.
Using the filtering approach, the matrix-vector operation are also
linear in complexity.
In terms of run-time, they represent the
costliest step so minimising the number of times that we are going to
perform them will gives us the best performance for our algorithm.
We remind the reader that the expression of the gradient used at an iteration is:
\begin{equation}
\label{eq:LRQP-grad}
\nabla S_{\text{cvx}}(\mathbf{y}) = (\bm{\phi} - \mathbf{d}) + 2 (\bm{\Psi} + \mathbf{D}) \mathbf{y}.
\end{equation}
so by keeping intermediary results of the gradient's computation, we don't need to compute $(\bm{\Psi} + \mathbf{D}) \mathbf{y}$, having already performed this operation once.
The other matrix-vector product that is necessary for obtaining the optimal step-size is $(\bm{\Psi} + \mathbf{D}) \mathbf{s}$.
During the first iteration, we will need to compute is using filter-based methods, which means using them twice in the same iteration. However, this computation can be reused.
The update rules that we follow are:
\begin{equation}
\label{eq:conditional-upd-rule}
\mathbf{y}^{t+1} = \mathbf{y}^{t} + \alpha (\mathbf{s} - \mathbf{y}^{t}).
\end{equation}
At the following iteration, to obtain the gradient, we will need to compute:
\begin{equation}
\begin{split}
(\bm{\Psi} + \mathbf{D}) \mathbf{y}^{t+1} &= (\bm{\Psi} + \mathbf{D}) (\mathbf{y}^{t} + \alpha (\mathbf{s} - \mathbf{y}^{t}),\\
&= (1 - \alpha) (\bm{\Psi} + \mathbf{D}) \mathbf{y}^{t} + \alpha (\bm{\Psi} + \mathbf{D}) \mathbf{s}.
\end{split}
\end{equation}
All the matrix-vector product of this equation have already been computed.
This means that no call to the filter-based method will be required.

At each iteration, we will only need to perform the expensive matrix-vector products on the conditional gradient.
Using linearity and keeping track of our previous computations, we can then obtain all the other terms that we need.

\section{Convex problem in the restricted DC relaxation}
\label{supp-dcneg}
Two difference-of-convex decompositions of the objective function are presented in the paper.
The first one is based on diagonally dominant matrices to ensure convexity and would be applicable to any QP objective function.
However, using this decomposition, the terms involving the pixel-compatibility function, and therefore requiring filter-based convolutions, need to be computed several times per CCCP iteration.

On the other hand, in the case of negative semi-definite compatibility functions, a decomposition suited to the structure of the problem is available.
Using this decomposition, similar to the one proposed by Kr\"{a}henb\"{u}hl \cite{denseccp_kra}, the convex problem to solve CCCP will be the following:
\begin{equation}
\label{eq:convexpb-concavecccp}
\begin{split}
\min &\quad (\bm{\phi}^{T} - \mathbf{g}^{T}) \mathbf{y} - \mathbf{y}^{T} (\bm{\mu} \otimes \mathbf{I}_{N}) \mathbf{y},\\
\text{s.t. } & \quad \mathbf{y} \in \mathcal{M}.
\end{split}
\end{equation}
The filter-method has been used to compute the gradient of the concave part $\mathbf{g}$. The Kronecker product with the identity matrix will make this problem completely de-correlated between pixels.
This means that instead of solving one problem involving $N \times L$ variables, we will have to solve $N$ problems of $L$ variables, which is much faster. The problem to solve for each pixel are, using the $a$ subscript to refer to the subset of the vector elements that correspond to the random variable $a$:
\begin{equation}
\label{eq:convexpb-onevar}
\begin{split}
\min &\quad (\bm{\phi}_{a}^{T} - \mathbf{g}_{a}^{T}) \mathbf{y}_{a} - \mathbf{y}_{a}^{T} \bm{\mu}  \mathbf{y}_{a},\\
\text{s.t. } & \quad \mathbf{y}_{a} \geq 0\\
& \quad \mathbf{y}_{a}^{T} \mathds{1} = 1.
\end{split}
\end{equation}
These problems can also be solved using the Frank-Wolfe algorithm, with efficient conditional gradient computation and optimal step size.
The only difference is that in that case, no filter-based method will need to be used for computation.
CCCP on this DC relaxation will therefore be much faster than on the generic case, an improvement gained at the cost of generality.

We also remark that the guarantees of CCCP to provide better results at each iteration does not require to solve the convex problem exactly. It is sufficient to obtain a value of the convex problem lower that the initial estimate. Therefore, the inference may eventually be sped-up by solving the convex problem approximately instead of reaching the optimal solution.

\section{LP objective reformulation}
\label{supp-lp-reform}
This section presents the reformulation of the pairwise part of the LP objective.
We first introduce the following equality:
\begin{equation}
\sum_{a} \sum_{b > a} K_{a,b} y_{b}(i) = \sum_{a} \sum_{b < a} K_{a,b} y_{a}(i),
\end{equation}
It comes from the symmetry of $\mathbf{K}$.

Using the above formula, considering the reordering has already been done, we can rewrite the pairwise term of (18) as:
\begin{equation}
\begin{split}
  & \sum_{a} \sum_{b \neq a} \sum_{i} K_{a,b} \frac{|y_{a}(i) - y_{b}(j)|}{2}, \\
  =& \sum_{i} \sum_{a} \sum_{b > a} K_{a,b} \frac{y_{a}(i) - y_{b}(i)}{2} - \sum_{i} \sum_{a} \sum_{b < a} K_{a,b} \frac{y_{a}(i) - y_{b}(i)}{2}, \\
  =& \sum_{i} \sum_{a} \sum_{b > a} K_{a,b} y_{a}(i) - \sum_{i} \sum_{a} \sum_{b < a} K_{a,b} y_{a}(i).
\end{split}
\end{equation}
It is important to note that in these equations, the ordering between $a$ and $b$ used in the summations is dependent on the considered label $i$.

\section{LP Divide and conquer}
\label{supp-lp-dc}
We are going to present an algorithm to efficiently compute the following:
\begin{equation}
  \forall k \quad \sum_{j > k} K_{k,j},
\end{equation}
for $j$ and $k$ being between $1$ and $N$.
For the sake of simplicity, we are going to consider $N$ as being even. The odd case is very similar.
Considering $h = N/2$, we can rewrite the original sum as:

\begin{equation}
  \begin{split}
    \sum_{j > k} K_{k,j} =&
      \begin{cases}
          \hfill \sum_{j > k} K_{k,j} \hfill & \text{ if } k > h \\
          \hfill \sum_{j > k} K_{k,j} \hfill & \text{ if } k \leq h \\
      \end{cases} \\
    =&
      \begin{cases}
          \hfill \sum_{j > k} K_{k,j} \hfill & \text{ if } k > h \\
          \hfill \sum_{\substack{j > k\\j \leq h}} K_{k,j} + \sum_{\substack{j > k\\j > h}} K_{k,j}\hfill & \text{ if } k \leq h \\
      \end{cases} \\
    =&
      \begin{cases}
          \hfill \underbrace{\sum_{j > k} K_{k,j}}_{A} \hfill & \text{ if } k > h \\
          \hfill \underbrace{\sum_{\substack{j > k\\j \leq h}} K_{k,j}}_{B} + \underbrace{\sum_{\substack{j > h}} K_{k,j}}_{C} \hfill & \text{ if } k \leq h \\
      \end{cases} \\
  \end{split}
\end{equation}

We can see that both $A$ and $B$ corresponds to the cases where respectively $k,j > h$ and $k,j \leq h$. These two elements can be obtained by recursion using sub-matrices of $\mathbf{K}$ which have half the size of the current size of the problem.
To compute the $C$ part, we consider the following variable:
\begin{equation}
  v_j =
      \begin{cases}
          \hfill 0 \hfill & \text{ if } j \leq h \\
          \hfill 1 \hfill & \text{ if } j > h \\
      \end{cases}
\end{equation}
We can now rewrite the $C$ part as $\sum_{\substack{j}} K_{k,j} v_j$. We can compute this sum efficiently for all $k$ using the filter-based method. Since this term contribute to the original sum only when $k \leq h$, we will only consider a subset of the output from the filter based method.

So we have a recursive algorithm that will have a depth of $log( N )$ and for which all level takes $\mathcal{O}( N )$ to compute. We can use it to compute the requested sum $\forall k$ in $\mathcal{O}(N log( N ))$.

\section{LP generalisation beyond Potts models}
\label{supp-genpairwise}
In this section, we consider the case where the label compatibility $\mu(x_{a}, x_{b})$ is any semi-metric. We recall that $\mu(\cdot, \cdot)$ is a semi-metric if and only if $d(i,i)=0, \forall i$ and $d(i,j)=d(j,i)>0, \forall i \neq j$.

To solve this problem, we are going to reduce the semi-metric labelling problem to a \textit{r-hierarchically well-separated tree} (r-HST) labelling problem that we can then reduce to a uniform labelling problem.

As described in~\cite{kumar2009map}, an r-HST metric \cite{bartal98} $d^t(\cdot, \cdot)$ is specified by a rooted tree whose edge lengths are non-negative and satisfy the following properties: (i) the edge lengths from any node to all of its children are the same; and (ii) the edge lengths along any path from the root to a leaf decrease by a factor of at least $r > 1$.
Given such a tree, known as r-HST, the distance $d^t(i, j)$ is the sum of the edge lengths on the
unique path between them

\subsection{Approximate the semi-metric with r-HST metric}
Fakcharoenphol et al.~\cite{Fakcharoenphol2003} present an algorithm to get in polynomial time a mixture of r-HST that approximate any semi-metric using a fixed number of trees. This algorithm generate a collection $\mathcal{D} = {d^t( \cdot , \cdot ), t=1,...,n}$ where each $d^t$ is a r-HST metric.

Since this mixture of r-HST metric is an approximation to the original metric, we can approximate the original labelling problem by solving the labelling problem on each of these r-HST metrics and combining them with the method presented in~\cite{kumar2009map}.
We are now going to present an efficient algorithm to solve the problem on an r-HST.
This algorithm can then be used to solve the problem in the semi-metric case.

\subsection{Solve the r-HST labelling problem}
We now consider a given r-HST metric $d^t$ and we note $\mathcal{T}$ all the sub-trees corresponding to this metric and we will use $T$ as one of these sub-trees.
This problem has been formulated by Kleinberg and Tardos \cite{kleinbergTardos2002}, but we are not using their method to solve it because the density of our CRF makes their method unfeasible. We are going to solve their original LP directly:

\begin{equation}
\label{eq:rHST-LP}
    \begin{split}
        \min \quad& \sum_{a} \sum_{i} \phi_{a}(i) y_{a}(i) + \sum_{a,b \neq a} \sum_{T} K_{a,b} c_{T} \frac{|y_{a}(T) - y_{b}(T)|}{2} \\
        \text{such that} \quad& \sum_{i} y_{a}(i) = 1 \quad \forall a \\
        & y_{a}(T) = \sum_{i \in L(T)} y_{a}(i) \quad \forall a \forall T \\
        & y_{a}(i) \in \{0,1\} \quad \forall a,i \\
    \end{split}
\end{equation}
Where $L(T)$ is the set of all labels associated with a sub-tree $T$.

This problem is the same the Potts model case except for two points:
\begin{itemize}
    \item In the pairwise term, the labels have been replaced by sub-trees. Since the assignment of the labels to the trees does not change, this won't prevent us from using the same method to compute the gradient.
    \item There is a factor $c_{T}$ corresponding to the weights in the tree. This does not prevents us from computing this efficiently since we can move this out of the inner loop with the summation over the trees.
\end{itemize}

Using the same trick where we sort the $y_{a}(T)$ for all $T$, we can rewrite the pairwise part of the above problem and compute its sub-gradient. Using the fact that $\frac{\partial y_a(T)}{\partial y_{c}(k)}$ is $0$ if $a \neq c$ or $k \not\in L(T)$ and $1$ otherwise and noting $T_k$ all the sub-trees that contains $k$ as one of their label.
\begin{equation}
    \begin{split}
        & \frac{\partial}{\partial y_{c,k}} (\sum_{T} \sum_{a,b \neq a} K_{a,b} c_{T} \frac{|y_{a}(T) - y_{b}(T)|}{2}) \\
        =& \frac{\partial}{\partial y_{c,k}} (\sum_{T} \sum_{a,b} K_{a,b} c_{T} \frac{y_{a}(T) - y_{b}(T)}{2}) - 2 \sum_{T} \sum_{a,b < a} K_{a,b} c_{T} \frac{y_{a}(T) - y_{b}(T)}{2})\\
        =& \sum_{T_k} \sum_{b} c_{T_k} \frac{K_{c,b}}{2} - \sum_{T_k} \sum_{a} c_{T_k} \frac{K_{a,c}}{2} - 2 \sum_{T_k} \sum_{b < c} c_{T_k} \frac{K_{c,b}}{2} + 2 \sum_{T_k} \sum_{a > c} c_{T_k} \frac{K_{a,c}}{2} \\
        =& - \sum_{T_k} \sum_{a < c} c_{T_k} K_{a,c} + \sum_{T_k} \sum_{a > c} c_{T_k} K_{a,c} \\
        =& \sum_{T_k} c_{T_k} (\sum_{a > c} K_{a,c} - \sum_{a < c} K_{a,c})
    \end{split}
\end{equation}
Since the sorting is done for each sub-tree $T$, the sums where we consider relative ordering of the indices cannot be switched and thus we cannot simplify this expression further.

We present a method to solve this problem in Algorithm~\ref{alg:rHST}.
State~\ref{eq:T:grad_init} initialise the subgradient with the unaries.
The loop at State~\ref{eq:T:subtree_loop} is used to compute the participation of each subtree to the total subgradient.
To do so, we first precompute the $y_{a}(T)$ terms for all pixel in the loop starting at State~\ref{eq:T:precomp}.
We then sort these $y_{a}(T)$ in State~\ref{eq:T:sort} and reorder the $\mathbf{K}$ matrix accordingly in State~\ref{eq:T:reorder}.
Using this, we can use the divide and conquer approach from the Potts model section to compute the participation of this tree to the subgradient for each pixel using the formula from State~\ref{eq:T:tree_grad}.
The vector $\mathbf{gt}$ is of size $N$ and contains one value per pixel.
We can now update the subgradient with the partial one on this tree for all labels associated with this tree.
The notation $\mathbf{g_{\cdot, k}}$ corresponds to a single row of the gradient matrix.
This is done in the for loop starting at State~\ref{eq:T:grad_accum}.
We can perform one step of subgradient descent in State~\ref{eq:T:descent}.
Finally we need to project the new point on the feasible set at State~\ref{eq:T:proj}

\begin{algorithm}
  \begin{algorithmic}[1]
    \State Get $\mathbf{y}^{0}$
    \While{not converged}
      \State Initialise the subgradient $\mathbf{g} = \bm{\phi}$ \label{eq:T:grad_init}
      \For{all subtree $T$} \label{eq:T:subtree_loop}
        \For{all pixel $a$} \label{eq:T:precomp}
            \State $y_{a}(T) = \sum_{i \in L(T)} y_{a}(i)$
        \EndFor
        \State Sort $y_{a}(T)$ \label{eq:T:sort}
        \State Reorder $\mathbf{K}$ \label{eq:T:reorder}
        \State Gradient for this subtree $gt_{c} = c_{T} (\sum_{a > c} K_{a,c} - \sum_{a < c} K_{a,c})$ \label{eq:T:tree_grad}
        \For{all label $k$} \label{eq:T:grad_accum}
          \If{$k \in T$}
            \State Update $\mathbf{g_{\cdot, k}} \mathrel{+}= \mathbf{gt}$
          \EndIf
        \EndFor
      \EndFor
      \State $ \mathbf{y}^{t+1} = \mathbf{y}^{t} - \beta \mathbf{g} $ \label{eq:T:descent}
      \State Project $\mathbf{y}^{t+1}$ such that it is a feasible point \label{eq:T:proj}
    \EndWhile
  \end{algorithmic}
  \caption{\label{alg:rHST} r-HST labelling problem}
\end{algorithm}

\FloatBarrier

\section{Model used in the experiments section}
\label{supp-compat-function}
All the results that we presented in the paper are valid for pixel compatibility functions composed of a mixture of Gaussian kernels:
\begin{equation}
\label{eq:pairwise}
\sum_{m} w^{(m)} k(\mathbf{f}^{(m)}_{a}, \mathbf{f}^{(m)}_{b}).
\end{equation}

In practice, for our experiments, we use the same form as Kr\"{a}henb\"{u}hl and Koltun~\cite{densecrf-kra}.
It is a mixture of two gaussian kernels defined using the position vectors $\mathbf{p}_{a}$ and $\mathbf{p}_{b}$, and the colour vectors $\mathbf{I}_{a}$ and $\mathbf{I}_{b}$ associated with each  pixel $a$ and $b$.
The complete formula is the following:
\begin{equation}
K_{a,b} = w^{(1)} \exp\left(-\frac{|\mathbf{p}_{a} - \mathbf{p}_{b}|^{2}}{\sigma_{1}} \right) + w^{(2)} \exp\left( -\frac{|\mathbf{p}_{a} - \mathbf{p}_{b}|^{2}}{\sigma_{2,\text{spc}}} - \frac{\mathbf{|\mathbf{I}_{a} - \mathbf{I}_{b}|^{2}}}{\sigma_{2,\text{col}}} \right).
\end{equation}
We note that this pixel compatibility function contains 5 learnable parameters \mbox{$w^{(1)}, \sigma_{1}, w^{(2)}, \sigma_{2,\text{spc}}, \sigma_{2,\text{col}}$}.

\section{More results on stereo}
\label{supp-stereo-res}
\begin{figure}[]
\captionsetup{font=footnotesize}
  \centering
\scalebox{0.9}{%
  \begin{tabular}{cccc}
  Left Image & \textbf{MF} & $\mathbf{DC_{neg}}$ & $\mathbf{QP_{cvx}}$ \\
  \includegraphics[width=0.2\columnwidth]{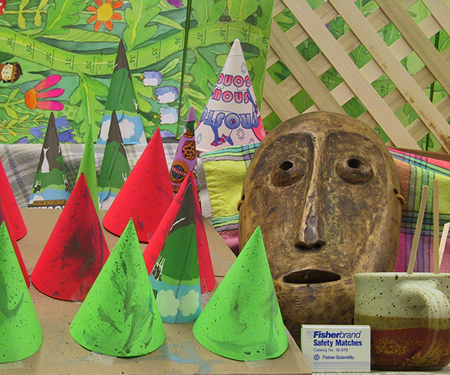} &
  \includegraphics[width=0.2\columnwidth]{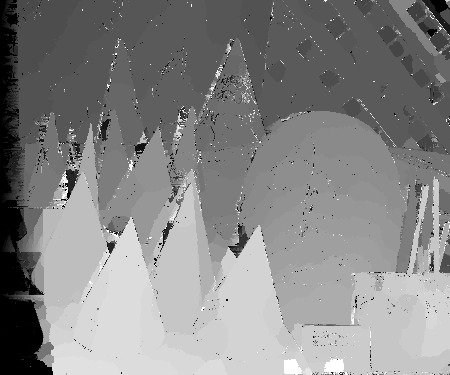} &
  \includegraphics[width=0.2\columnwidth]{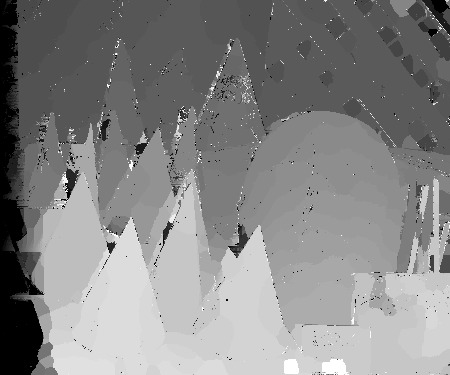} &
  \includegraphics[width=0.2\columnwidth]{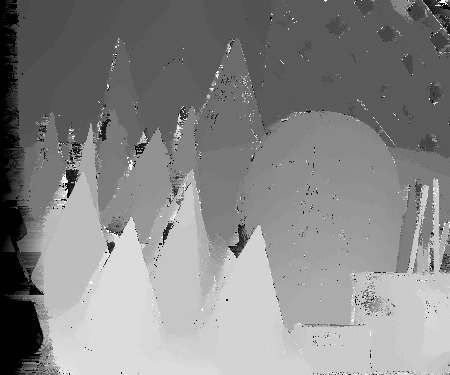} \\
  \includegraphics[width=0.2\columnwidth]{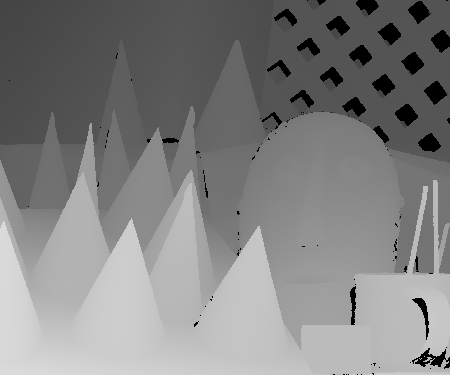} &
  \includegraphics[width=0.2\columnwidth]{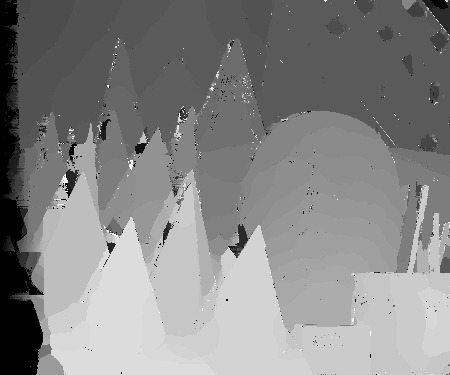} &
  \includegraphics[width=0.2\columnwidth]{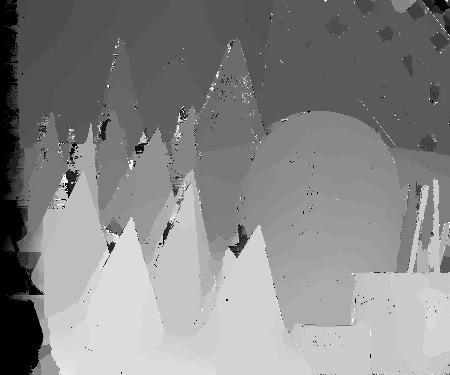} &
  \includegraphics[width=0.2\columnwidth]{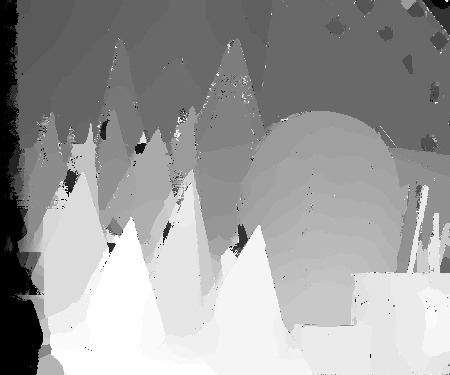} \\
  Ground Truth & $\mathbf{QP-DC_{neg}}$ & $\mathbf{QP-DC_{neg}}$ & $\mathbf{QP-DC_{neg}-LP}$ \\
  \end{tabular}
}
  \caption{Stereo matching results on the Cones image and corresponding timings. Here again, we can see that the \textbf{MF} solution is significantly better than the \textbf{MF5} solution. The continuous relaxations improve even further by reducing the number of artifacts.}
  \label{fig:cones_stereo}
\end{figure}

In the following Figures~\ref{fig:tsu_stereo}, \ref{fig:ven_stereo} and \ref{fig:cones_stereo}, we observe that the continuous relaxations give consistently better results.
We can also note that even though it runs only for a few iterations, the \textbf{LP} improves the visual quality of the solution.

\begin{figure}
  \begin{subfigure}{0.6\textwidth}
    \includegraphics[width=\columnwidth]{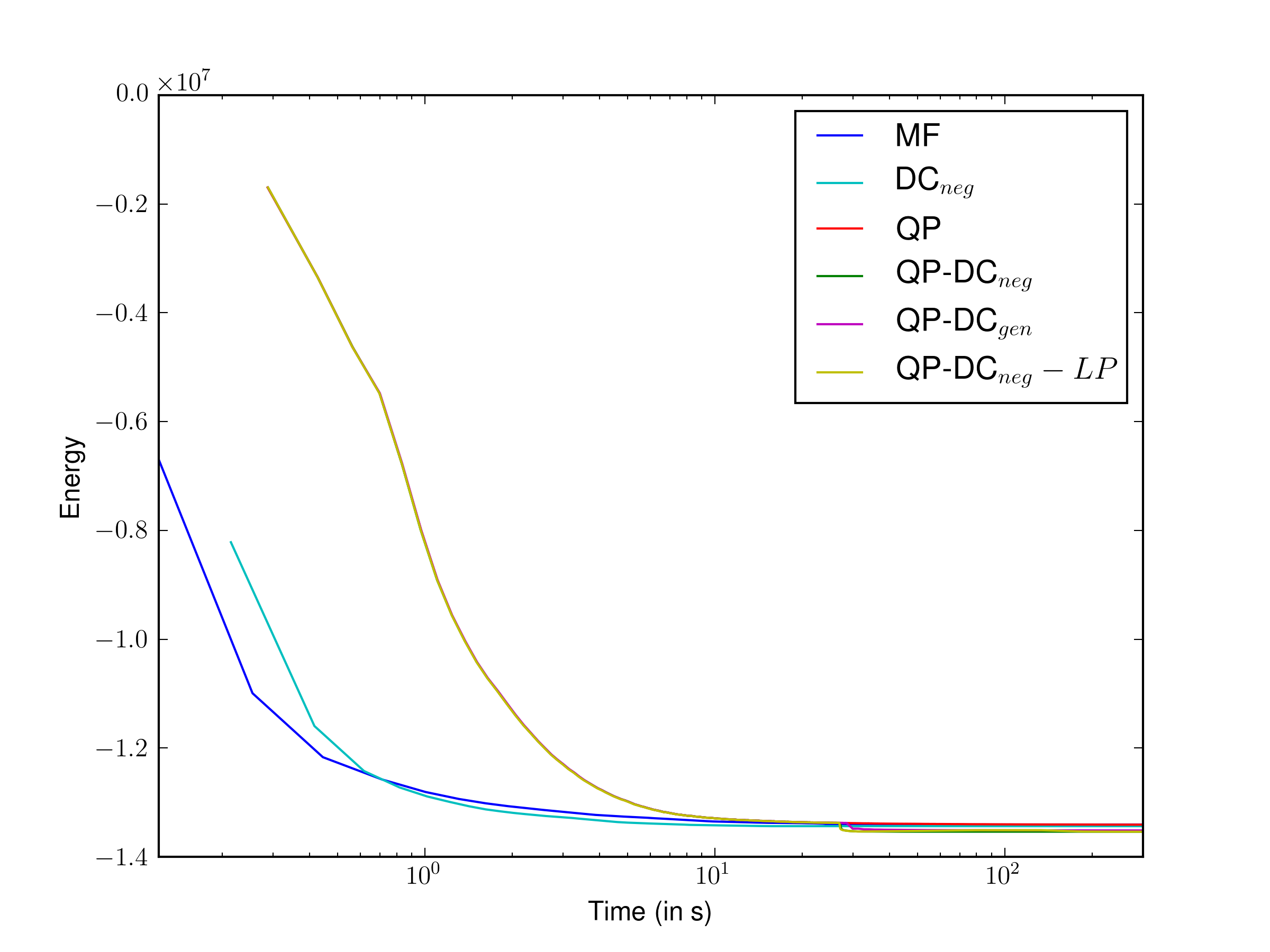}
  \end{subfigure}%
  \begin{subtable}{0.4\textwidth}
    \centering
  \begin{tabular}{|c|c|}
    \hline
    \textbf{Method} & \textbf{Final energy}\\
    \hline
    MF & -1.341e+07\\
    DC$_{neg}$ & -1.344e+07\\
    QP & -1.341e+07 \\
    QP-DC$_{neg}$ & \textbf{-1.354e+07} \\
    QP-DC$_{gen}$ & -1.352e+07 \\
    QP-DC$_{neg}$-LP & \textbf{-1.354e+07}\\
    \hline
  \end{tabular}
\end{subtable}
\caption{Evolution of achieved energies as a function of time on a stereo matching problem (Venus Image).}
\end{figure}

\begin{figure}
  \begin{subfigure}{0.6\textwidth}
    \includegraphics[width=\columnwidth]{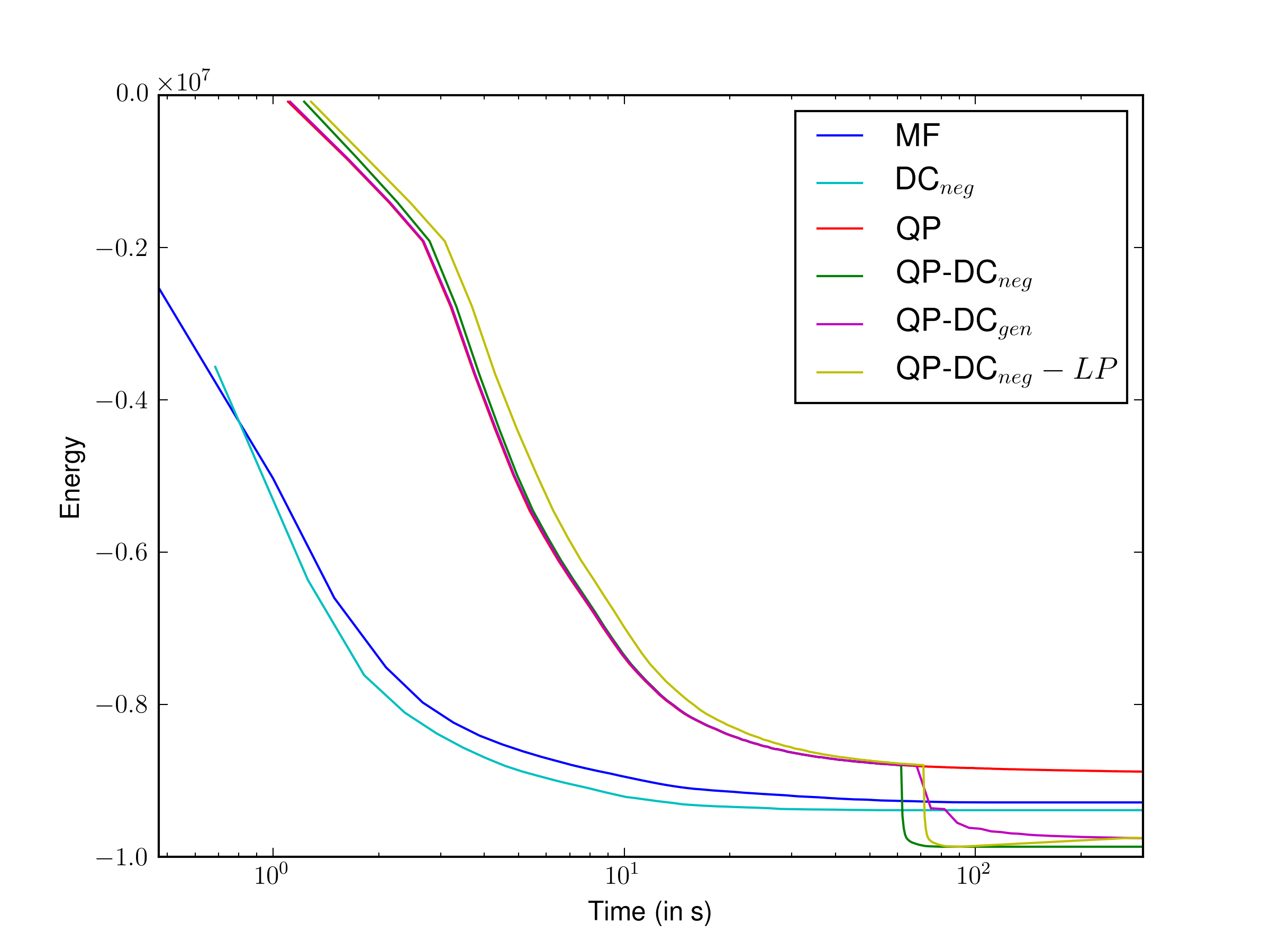}
  \end{subfigure}%
  \begin{subtable}{0.4\textwidth}
    \centering
  \begin{tabular}{|c|c|}
    \hline
    \textbf{Method} & \textbf{Final energy}\\
    \hline
    MF & -9.286e+06\\
    DC$_{neg}$ & -9.388e+06\\
    QP & -8.881e+06 \\
    QP-DC$_{neg}$ & \textbf{-9.868e+06} \\
    QP-DC$_{gen}$ & -9.757e+06 \\
    QP-DC$_{neg}$-LP & -9.758e+06\\
    \hline
  \end{tabular}
\end{subtable}
\caption{Evolution of achieved energies as a function of time on a stereo matching problem (Cones Image). Note that the LP in that case is not improving the results.}
\end{figure}

\begin{figure}
  \begin{subfigure}{0.6\textwidth}
    \includegraphics[width=\columnwidth]{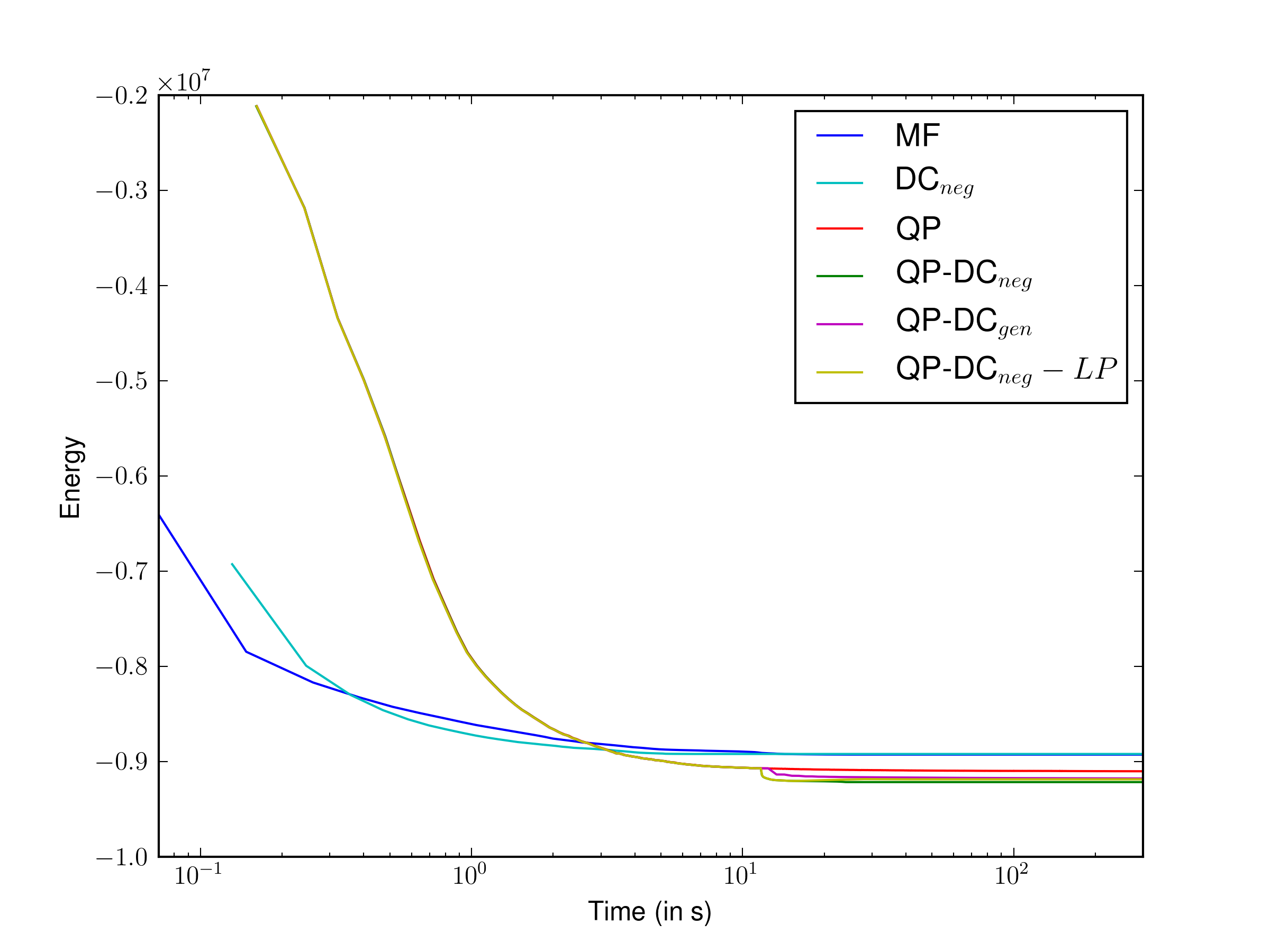}
  \end{subfigure}%
  \begin{subtable}{0.4\textwidth}
    \centering
  \begin{tabular}{|c|c|}
    \hline
    \textbf{Method} & \textbf{Final energy}\\
    \hline
    MF & -8.927e+06\\
    DC$_{neg}$ & -8.920e+06\\
    QP & -9.101e+06 \\
    QP-DC$_{neg}$ & \textbf{-9.215e+06} \\
    QP-DC$_{gen}$ & -9.177e+06 \\
    QP-DC$_{neg}$-LP & -9.186e+06\\
    \hline
  \end{tabular}
\end{subtable}
\caption{Evolution of achieved energies as a function of time on a stereo matching problem (Tsukuba Image). Note that the LP in that case is not improving the results.}
\end{figure}

\begin{figure}[]
\captionsetup{font=footnotesize}
  \centering
\scalebox{0.9}{%
  \begin{tabular}{cccc}
  Left Image & \textbf{MF} & $\mathbf{DC_{neg}}$ & $\mathbf{QP_{cvx}}$ \\
  \includegraphics[width=0.2\columnwidth]{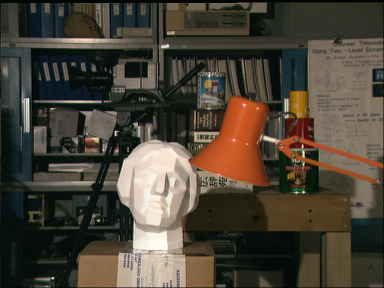} &
  \includegraphics[width=0.2\columnwidth]{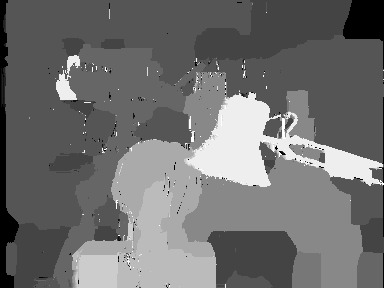} &
  \includegraphics[width=0.2\columnwidth]{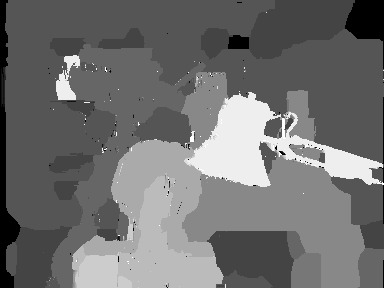} &
  \includegraphics[width=0.2\columnwidth]{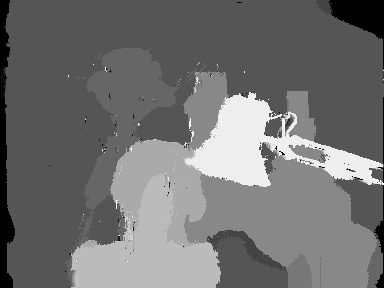} \\
  \includegraphics[width=0.2\columnwidth]{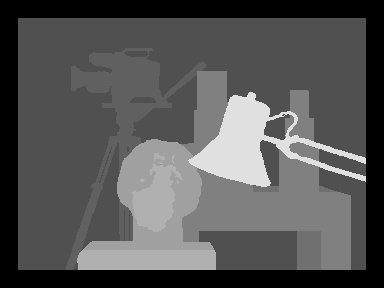} &
  \includegraphics[width=0.2\columnwidth]{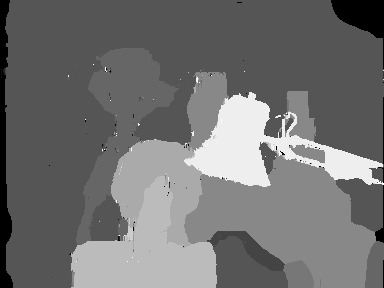} &
  \includegraphics[width=0.2\columnwidth]{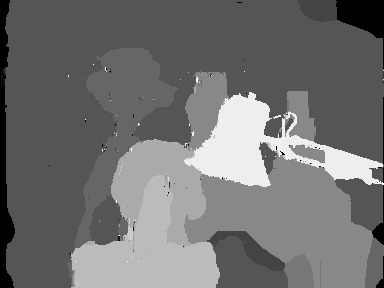} &
  \includegraphics[width=0.2\columnwidth]{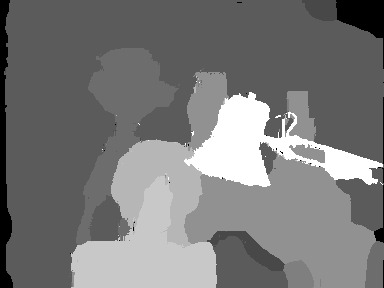} \\
  Ground Truth & $\mathbf{QP-DC_{neg}}$ & $\mathbf{QP-DC_{neg}}$ & $\mathbf{QP-DC_{neg}-LP}$ \\
  \end{tabular}
}
  \caption{Stereo matching results on the Tsukuba image and corresponding timings. We see that the LP method allows to improve the smoothness from its initialisation $\mathbf{DC_{neg}}$. For this set of parameters, the mean-field methods performs poorly.}
  \label{fig:tsu_stereo}
\end{figure}

\begin{figure}[]
\captionsetup{font=footnotesize}
  \centering
\scalebox{0.9}{%
  \begin{tabular}{cccc}
  Left Image & \textbf{MF} & $\mathbf{DC_{neg}}$ & $\mathbf{QP_{cvx}}$ \\
  \includegraphics[width=0.2\columnwidth]{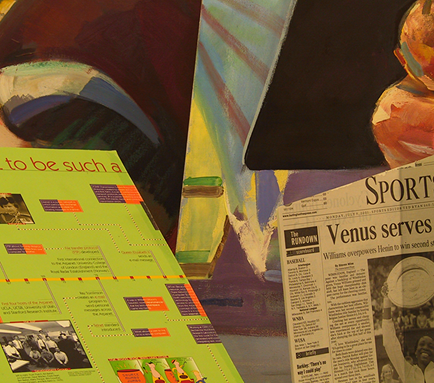} &
  \includegraphics[width=0.2\columnwidth]{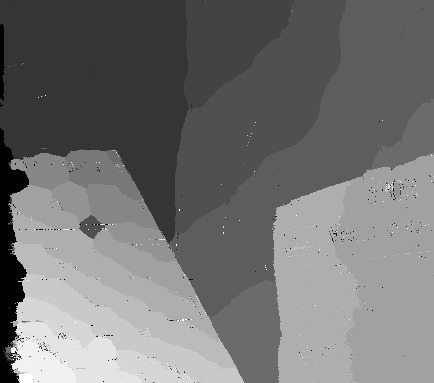} &
  \includegraphics[width=0.2\columnwidth]{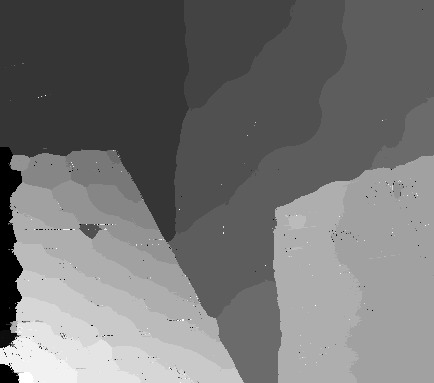} &
  \includegraphics[width=0.2\columnwidth]{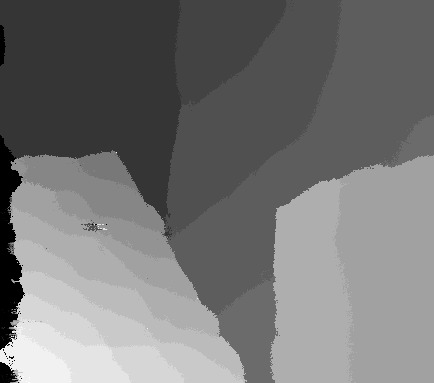} \\
  \includegraphics[width=0.2\columnwidth]{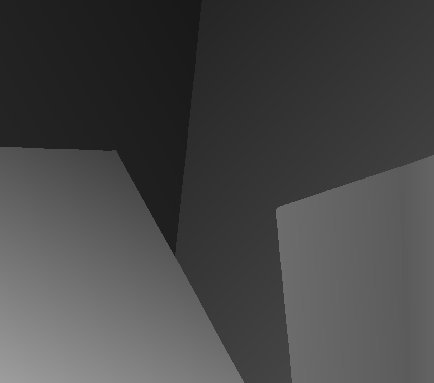} &
  \includegraphics[width=0.2\columnwidth]{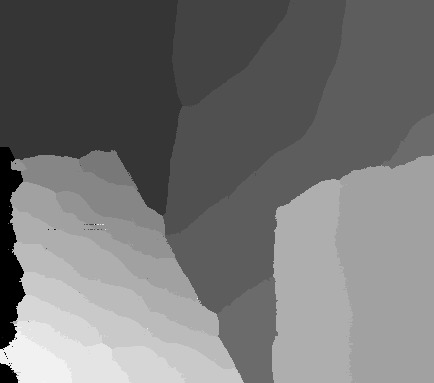} &
  \includegraphics[width=0.2\columnwidth]{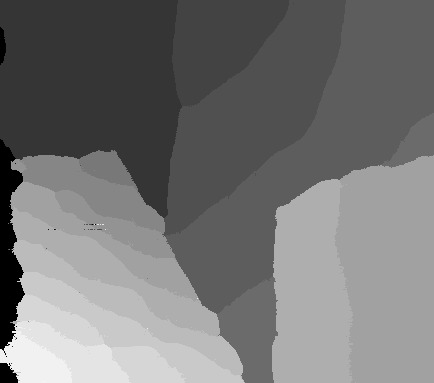} &
  \includegraphics[width=0.2\columnwidth]{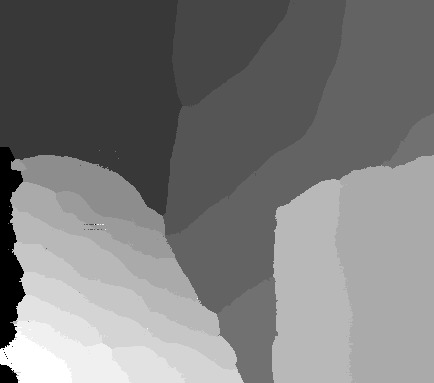} \\
  Ground Truth & $\mathbf{QP-DC_{neg}}$ & $\mathbf{QP-DC_{neg}}$ & $\mathbf{QP-DC_{neg}-LP}$ \\
  \end{tabular}
}
  \caption{Stereo matching results on the Venus image. Note that the smoothness of the reconstructions improves with methods reaching lower energies. The LP result does not artefacts anymore, only the non contiguous borders due to the Potts model.}
  \label{fig:ven_stereo}
\end{figure}

\FloatBarrier
\section{More results on segmentation}
\label{supp-seg-nrj}
Additional results for parameters cross-validated for \textbf{MF} are presented in Table~\ref{tab:nrj_on_mf}.

We see that in this case where the parameters are tuned for \textbf{MF}, the continuous relaxations still reach lower energy than the mean-field approaches on average.
Furthermore, we observe that in almost all images, the energy is strictly lower than the one provided by the mean-field methods.
However, with the parameters that were tuned for \textbf{MF} using cross-validation, we note that the segmentation performance is poor compared to mean-field approaches, and that the improved energy minimisation does not translate to better segmentation.

\begin{table}[h]
\captionsetup{font=footnotesize}
\renewcommand{\arraystretch}{1.1}
\setlength{\tabcolsep}{0.7ex}
\centering
\begin{tabular}{@{}c@{\hspace*{2ex}}ccccccc@{\hspace*{3ex}}c@{\hspace*{3ex}}cc@{}}
\toprule
 & Unary & \textbf{MF5} & \textbf{MF} & $\mathbf{QP_{cvx}}$ & $\mathbf{DC_{gen}}$ & $\mathbf{DC_{neg}}$ & \textbf{LP} & \textbf{Avg. E} & \textbf{Acc} & \textbf{IoU}\\
\midrule
Unary & - & 0 & 0 & 0 & 0 & 0 & 0 & 0 & 79.04 & 27.43\\
\cmidrule(r{2ex}){2-8}\cmidrule(r{2ex}){9-9}\cmidrule(){10-11}
\textbf{MF5} & 99 & - & 0 & 1 & 0 & 1 & 1 & -8.37e5 & 80.42 & 28.66\\
\textbf{MF} & 99 & 93 & - & 4 & 2 & 3 & 3 & -1.19e6 & \textbf{80.95} & \textbf{28.86}\\
\cmidrule(r{2ex}){2-8}\cmidrule(r{2ex}){9-9}\cmidrule(){10-11}
$\mathbf{QP_{cvx}}$ & 99 & 93 & 86 & - & 0 & 4 & 2 & -1.66e6 & 77.75 & 14.94\\
$\mathbf{DC_{gen}}$ & 99 & 94 & 88 & 32 & - & 29 & 29 &\textbf{-1.68e6} & 77.76 & 14.96\\
$\mathbf{DC_{neg}}$ & 99 & 93 & 87 & 27 & 2 & - & 12 & -1.67e6 & 77.76 & 14.91\\
\textbf{LP} & 99 & 93 & 87 & 29 & 2 & 17 & - & -1.67e6 & 77.77 & 14.93\\
\bottomrule
\end{tabular}
\caption{Percentage of images obtaining strictly lower energy values. Average energy results over the test set and Segmentation performance. Higher percentage is better and lower energy is better. Higher accuracy and IoU are better. Continuous relaxations dominate mean-field approaches on almost all images and improve significantly more compared to the Unary baseline. Parameters tuned for \textbf{MF}.}
\label{tab:nrj_on_mf}
\end{table}

\FloatBarrier

\end{document}